\documentclass[10pt,twocolumn,letterpaper]{article}

 \usepackage[pagenumbers]{cvpr} 

\usepackage{multirow}

 \renewcommand{\paragraph}[1]{\vspace{.7em}\noindent\textbf{#1}}

\usepackage[outline]{contour}
\newcommand{\uparrowaligned}{\raisebox{0.15em}{\scalebox{0.75}{\contour{black}{$\uparrow$}}}}
\newcommand{\downarrowaligned}{\raisebox{0.15em}{\scalebox{0.75}{\contour{black}{$\downarrow$}}}}

\usepackage{colortbl}
\newcommand{\cellbest}{\cellcolor{red!23}}
\newcommand{\cellsecond}{\cellcolor{orange!26}}

\usepackage{multirow}
\usepackage{tabularx}
\newcolumntype{Y}{>{\centering\arraybackslash}X}
\newcolumntype{C}[1]{>{\centering}p{#1}}
\newcolumntype{Z}{>{\raggedleft\arraybackslash}X}

\definecolor{cvprblue}{rgb}{0.21,0.49,0.74}
\usepackage[pagebackref,breaklinks,colorlinks,allcolors=cvprblue]{hyperref}
\usepackage{adjustbox}
\newlength{\kubric}
\newlength{\kubricr}
\newlength{\kubricc}
\def\paperID{45028} 
\def\confName{CVPR}
\def\confYear{2026}

\title{Off The Grid: Detection of Primitives for Feed-Forward 3D Gaussian Splatting}

\author{Arthur Moreau \quad Richard Shaw \quad Michal Nazarczuk \quad Jisu Shin \quad
Thomas Tanay \\ Zhensong Zhang \quad Songcen Xu \quad Eduardo Pérez-Pellitero
\\ Huawei Noah's Ark Lab}

\begin{document}

\twocolumn[{%
\renewcommand\twocolumn[1][]{#1}%
\maketitle
\centering
\vspace{-2em}
\includegraphics[width=0.8\linewidth]{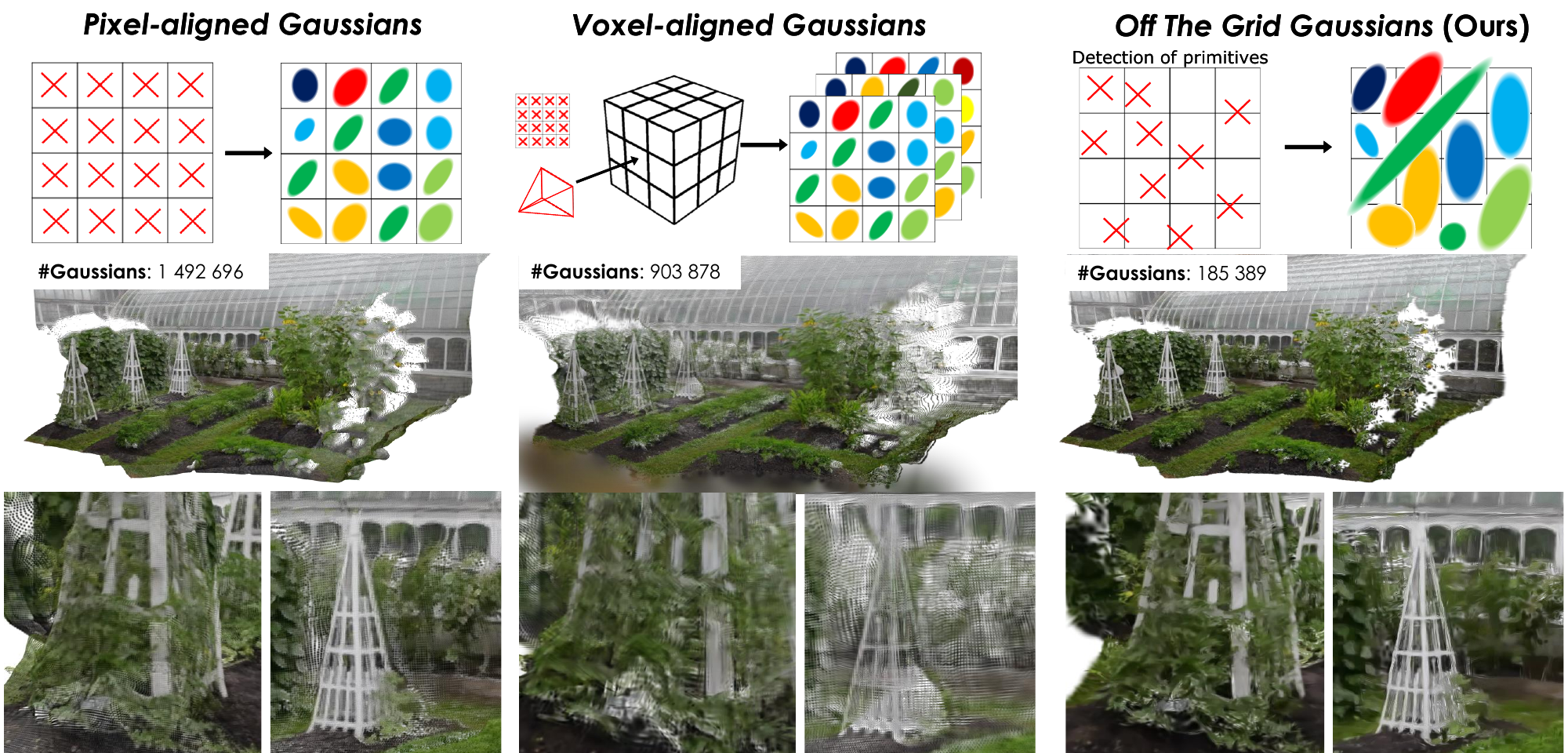}
\captionof{figure}{\textbf{3D Gaussians placement techniques.} Our model learns the position of primitives instead of using regular grids, representing the scene more accurately with fewer primitives. Voxel-aligned uses AnySplat~\cite{jiang2025anysplat} and Pixel-aligned is an ablated version of our model. 
\vspace{6pt}
}
\label{fig:teaser}
}]


\begin{abstract}
Feed-forward 3D Gaussian Splatting (3DGS) models enable real-time scene generation but are hindered by suboptimal pixel-aligned primitive placement, which relies on a dense, rigid grid that limits both quality and efficiency. We introduce a new feed-forward architecture that detects 3D Gaussian primitives at a sub-pixel level, replacing the pixel grid with an adaptive, ``Off-The-Grid" distribution. Inspired by keypoint detection, our decoder learns to locally distribute primitives across image patches. We also provide an Adaptive Density mechanism by assigning varying number of primitives per patch based on Shannon entropy. We combine the proposed decoder with a pre-trained 3D reconstruction backbone and train them end-to-end using photometric supervision without any 3D annotation. The resulting pose-free model generates photorealistic 3DGS scenes in seconds, achieving state-of-the-art novel view synthesis for feed-forward models. It outperforms competitors while using far fewer primitives, demonstrating a more accurate and efficient allocation that captures fine details and reduces artifacts. Project page:  \href{https://arthurmoreau.github.io/OffTheGrid/}{https://arthurmoreau.github.io/OffTheGrid/}.
\end{abstract}

\section{Introduction}
\label{sec:intro}

The recent introduction of 3D Gaussian Splatting~\cite{kerbl3Dgaussians} (3DGS) has marked a significant leap forward in the reconstruction of photorealistic 3D scenes. This point-based scene representation is highly efficient to render, enabling photorealistic interactive applications~\cite{Matsuki2024, moreau2024human, jiang2024hifi4g, shaw2026ico3d}, although it is generally slower to fit from a set of images than some of its predecessors\ \cite{mueller2022instant}. Starting from images, the standard pipeline first performs 3D reconstruction with Structure-from-Motion~\cite{schoenberger2016sfm} (SfM), and then optimizes Gaussian primitive parameters by rendering images in an iterative fashion. Following the initial optimization procedure, this process typically takes tens of minutes to hours. 

An alternative approach uses feed-forward models~\cite{charatan23pixelsplat,jiang2025anysplat} that predict 3D Gaussians directly from images through neural networks, enabling scene reconstruction in a single feed-forward step. Early research has primarily focused on developing \textit{pose-free} methods~\cite{ye2024no, huang2025no, kang2025selfsplat} that remove dependence on pre-computed camera poses. This task is challenging because the model needs to solve the 3D reconstruction problem before predicting photorealistic primitives. Less attention has been given to designing effective ways to accurately decode 3D Gaussian primitives from images to obtain higher-quality models. 

Existing works typically predict \textit{pixel-aligned Gaussians}~\cite{charatan23pixelsplat}, unprojecting one primitive per input pixel into the 3D scene using depth. With as many primitives as input pixels, these models are limited to operating at low resolutions with sparse image collections. Beyond primitive count, we question whether a regular grid-based distribution is optimal to obtain photorealistic models. As a comparison, optimization techniques for Gaussian Splatting use densification and pruning strategies to distribute primitives across the scene and ensure a good representation of high-frequency details. Current feed-forward models lack the ability to achieve this. Toward more accurate and scalable solutions, we argue that feed-forward models require more expressivity in the positioning of primitives.

To address this, we propose to adaptively control the allocation of primitives at 3 different levels. First, we \textit{detect} 3D Gaussian primitives in image patches at sub-pixel level, inspired by common keypoint detection techniques~\cite{detone2018superpoint}. Naturally, there is no actual ellipsoid primitive to be detected in the physical world. However, we see an opportunity to leverage 2D keypoints as a way to learn how to optimally distribute primitives across the image in a self-supervised manner. At a local level, the model can place primitive centers~\textit{off the grid} by extracting continuous 2D point coordinates from convolutional heatmaps~\cite{nibali2018numerical}.  Then, we define an adaptive density mechanism with a multi-density decoder, where high entropy patches are assigned more primitives than homogeneous ones. Finally, we learn per-Gaussian confidence values to aggregate primitives from different images, giving the model the ability to discard primitives when they are not needed. 

This decoding strategy is combined with a large 3D reconstruction model, VGGT~\cite{wang2025vggt}. Given the geometry predicted by this model, our decoder learns where and how to place Gaussians on this geometry to obtain a photorealistic model. By rendering the 3D Gaussians back to the input images, we define an end-to-end self-supervised loop that learns primitive detection and fine-tunes the reconstruction model without any annotation. We obtain a pose-free 3DGS model that can generate photorealistic reconstructions of any scene within seconds. We outperform state-of-the-art competitors on novel view synthesis while using 7 times fewer primitives than input pixels. We observe that our method fit details accurately, avoid floating artifacts, and allocate computational resources more effectively.

In summary, we introduce the following contributions:
\begin{enumerate}
    \item A solution to place primitives at a sub-pixel level, performing better than pixel and voxel aligned approaches,
    \item A multi-density adaptive mechanism that allocates primitives dynamically depending on image patch content,
    \item A pose-free feed-forward Gaussian Splatting model that outperforms existing methods on novel view synthesis.
\end{enumerate}

\section{Related Work}
\label{sec:related_work}

\paragraph{Neural Rendering of 3D Scenes.}
Neural Radiance Fields (NeRF) by Mildenhall \etal\cite{mildenhall2021nerf} introduced a new paradigm to fit 3D representations to 2D posed images through differentiable volumetric rendering coupled with stochastic gradient descent. 3D Gaussian Splatting (3DGS)~\cite{kerbl3Dgaussians} represented a breakthrough in rendering efficiency and overcame some of NeRF's inherent computational bottlenecks. 3DGS represents the world with a set of volumetric primitives shaped as 3D Gaussians. This set is initialized from a sparse point cloud, typically obtained via SfM~\cite{schoenberger2016sfm}, and then iteratively optimized to render the input images. During this process, the set of primitives is gradually pruned or densified, based on \eg photometric gradient magnitude\,\cite{kerbl3Dgaussians}, or other heuristics~\cite{rota2024revising, kim2024color, lyu2025resgs}, aiming to allocate primitives efficiently where representation capacity is needed. Despite notable acceleration techniques~\cite{Mallick2024taming3dgs,Matsuki2024,chen2025dashgaussian, meuleman2025fly}, scene-optimization methodologies often still require pre-computed 3D camera poses and scene point clouds. Accelerating the fitting stage, and defining improved strategies for the sampling and distribution of primitives during that process remain active topics of research\,\cite{deng2025,ren2025fastgs}.

\paragraph{Feed-forward 3DGS.}
An alternative to optimization is to train neural networks to efficiently predict the 3D Gaussian model in a single forward pass. This idea builds on generalizable view synthesis methods~\cite{tanay2024global,jin2025lvsm, jiang2025rayzer} that have shown great success in learning priors for rendering but generate novel views directly instead of a 3D model much faster to render. The pioneer work for Feed-forward 3DGS is PixelSplat~\cite{charatan23pixelsplat}, that uses \textit{pixel-aligned Gaussians}, \eg one primitive per input pixel. This design is popular in the prior art~\cite{szymanowicz2024splatter_image, gslrm2024, xu20254dgt, xu2024depthsplat, wang2024freesplat, wang2025freesplat++}, but the large number of primitives limits its application to low resolutions (typically $256 \times 256$) and sparse image collections. Moreover, areas observed in several images contain redundant primitives, leading to blurry renderings. 4DGT~\cite{xu20254dgt} shows that such a dense representation is not necessary, as many primitives can be pruned at test time. Alternatively, \textit{voxel-aligned Gaussians} have also been proposed~\cite{wang2025volsplat,jiang2025anysplat}, using unprojected features to predict one primitive per voxel to address the multi-view aggregation problem. We observe that these approaches exhibit noisy geometry, and the regular grid is often visible when zoomed in (see Figure~\ref{fig:extrap_anysplat}). MVSplat~\cite{chen2024mvsplat} uses a plane-sweeping cost volume which also discretizes 3D space in a regular grid. Both \textit{pixel-aligned} and \textit{voxel-aligned} strategies distribute primitives in a predefined regular structure, whereas the best optimization-based methods do not, limiting the expressivity of the model to place primitives optimally in the scene. Our work proposes \textit{Off-The-Grid Gaussians}, which are placed with sub-pixel precision and do not require explicit 3D aggregation.

\begin{figure*}[t]
\centering
\includegraphics[width=0.99\textwidth]{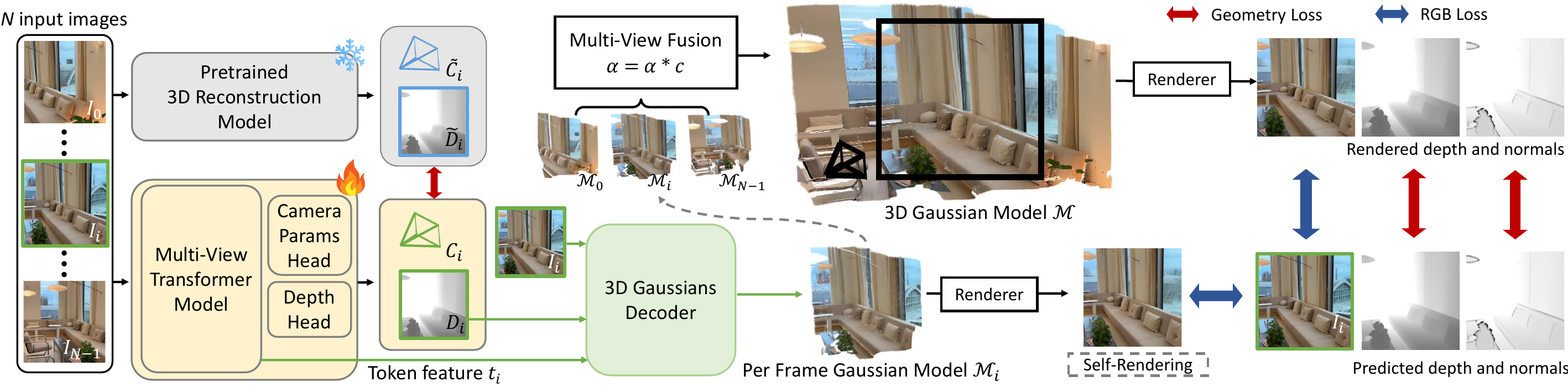}
\caption{\textbf{Overview of our pose-free 3DGS training framework.} We process depth and camera parameters from N input images with a large reconstruction model. Then, our 3D Gaussian decoder predicts primitives for each image, which are rendered and aggregated with other views. The pipeline is trained end-to-end to reconstruct input images, with geometry consistency and regularization losses.}
\label{fig:method}
\end{figure*}

\paragraph{Pose-free Methods.}
An important limitation in the scalability of Gaussian models is their dependence on pre-computed camera poses. One line of research has developed \textit{pose-free} approaches~\cite{fu2024colmap, fan2024instantsplat, jiang2024construct} that solve the camera pose estimation problem jointly with Gaussian model reconstruction. In the context of feed-forward methods, several works such as FLARE~\cite{zhang2025flare}, PF3Splat~\cite{hong2024pf3plat}, and VicaSplat~\cite{li2025vicasplat} have developed custom pipelines that first predict camera poses and then reconstruct the Gaussian primitives. However, SfM remains a notoriously difficult problem for learning-based methods, which are still outperformed by classical pipelines based on correspondences~\cite{schoenberger2016sfm}. This observation has been recently challenged by the emergence of 3D foundation models~\cite{wang2024dust3r, leroy2024grounding, wang2025vggt, wang2025pi3, keetha2025mapanything}, trained in a supervised manner on large-scale datasets to reconstruct 3D scene geometry from images, predicting camera parameters, depth maps, and/or point maps. These models offer a great opportunity to build pose-free feed-forward 3DGS methods by adding a decoder that predicts Gaussian primitives. NoPoSplat~\cite{ye2024no}, Splatt3R~\cite{smart2024splatt3r}, and SPFSplat~\cite{huang2025no} build on Mast3R~\cite{leroy2024grounding} and predict Gaussian centers as point maps, but process only image pairs and lack consistency with more images. AnySplat~\cite{jiang2025anysplat} fine-tunes VGGT~\cite{wang2025vggt} to predict 3D Gaussians, using pixel-aligned Gaussians aggregated in a voxel grid. We propose a pose-free method that also fine-tunes VGGT combined with a different decoder that leads to better visual quality and compactness.

\section{Method}
\label{sec:method}

We present a feed-forward neural network that predicts a 3DGS model from a set of $N$ unposed and uncalibrated images $I_{i<N}$ representing a static scene. An overview of our pipeline is shown in Figure~\ref{fig:method}.

\begin{figure*}[t]
\centering
\includegraphics[width=0.99\textwidth]{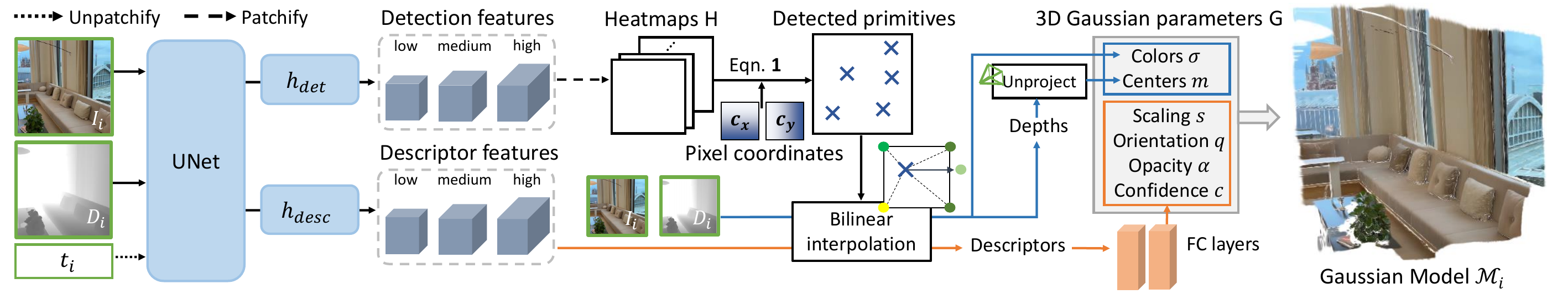}
\caption{\textbf{Overview of our 3D Gaussian decoder architecture.} Images, depth maps, and latent features are concatenated and fed to a U-Net CNN from which detection and description features are extracted. First, the position of detected primitives is determined from convolutional heatmaps. Then, image, depths and description features are bilinearly interpolated to decode Gaussian parameters through depth unprojection and MLP.}
\label{fig:decoder}
\end{figure*}

\subsection{Feed-forward 3D Reconstruction Backbone}

We use VGGT~\cite{wang2025vggt} as the backbone of our model. It is a large multi-view transformer that performs 3D reconstruction from unposed image collections in a single forward pass. Each image is first encoded into $14\times14$ patches through DINOv2~\cite{oquab2024dinov2}. Then, 24 transformer blocks alternate between global and frame-only attention. Output tokens $t_{i}$ are decoded into depth maps $D_{i}$ and extrinsics $[R_{i}|T_{i}]$ and intrinsics $K_{i}$ camera parameters. Instead of using the point map head, we compute 3D positions by combining the predicted depth and camera parameters. This model is trained to predict geometry but has no rendering ability. Similar to AnySplat~\cite{jiang2025anysplat}, we fine-tune it to predict 3D Gaussians. In contrast, we do not use a separate depth head for rendering. We fine-tune the backbone with 2 purposes: encode multi-view information useful for our task in the latent features and improve the geometry of the model from photometric supervision. Although we observe the 3D geometry provided by VGGT to be fairly accurate, predicted depth for background pixels is often inaccurately close, producing floating artifacts. Our method mitigates this issue by fine-tuning and generates 3D models with significantly cleaner geometry.

\subsection{3D Gaussian Decoder}
\label{sec:method_decoder}

We define a convolutional module, depicted in Figure~\ref{fig:decoder}, that operates after the decoding of camera parameters and depth maps. The objective of this decoder is to \textit{detect} 2D locations of primitives and to \textit{describe} each primitive to predict remaining parameters. It does not modify the 3D geometry provided by the backbone but places primitives on it. The decoder takes as input VGGT output tokens $(t_{i})$ but also input images $I_{i}$ and predicted depth maps $D_{i}$. Tokens are transformed to an image shape by \textit{unpatchifying}, i.e. project the 1D vectors to values of a $14\times14$ patch with 8 channels with a fully-connected layer and reshaping. Inputs are then concatenated on the channels dimension and fed to the decoder. As detection is essentially a low-level vision problem, we use a simple U-Net~\cite{ronneberger2015u} convolutional architecture. It outputs features with the same spatial dimension as inputs and 32 channels, from which detection and description features are extracted separately through heads $h_{det}$ and $h_{desc}$. 

\paragraph{2D Detection of Gaussian Primitives.}
Primitives are not attached to a pixel but to floating-point 2D coordinates in image space $(x,y)$. We use a heatmap approach to extract these coordinates in a differentiable manner. First, we reshape detection features back to $14\times14$ patches with $P$ channels. $P$ is the number of primitives per patch, i.e. each primitive is assigned a one-channel patch. We perform softmax over spatial dimensions to obtain a heatmap for each primitive, interpreted as the distribution of the primitive center position in the patch. By using tensors containing pixel coordinates $c_{x}$ and  $c_{y}$, the expectation of Gaussian positions $(x,y)$ can be simply computed by :

\begin{align}
    x = \sum_{i,j=0}^{P} c_{x}(i,j) h(i,j) && y = \sum_{i,j=0}^{P} c_{y}(i,j) h(i,j)
\end{align}

This operation can also be seen as a soft-argmax. Nibali et al.~\cite{nibali2018numerical} named it DSNT and showed improved performance on Human Pose Estimation. One main advantage is that keypoints are not limited to the pixel grid but are defined in the continuous 2D space. If we want a primitive to represent 2 neighboring pixels, the best Gaussian center position is between pixels. Our model can naturally achieve that by activating the 2 pixels equally in the heatmap. By performing this operation on all heatmaps of all patches, we obtain the full set of Gaussian centers.

\paragraph{Adaptive Density of Detection.} Instead of assigning a constant number of primitives to each image patch, we want to allocate more primitives to highly detailed areas. We define multiple levels of density with increasing numbers of Gaussians per patch. We follow APT~\cite{choudhury2025accelerating} and use Shannon entropy as a measure of patch compressibility. Similar to their work, we compute per-patch histograms of grayscale intensities $(p_{k})_{k \in [0,255]}$ and define per-patch entropy as $H = -\sum_{k=0}^{K-1} p_k \log_2(p_k + \epsilon)$. We assign the 55\% lowest entropy patches to `low density' (16 primitives), the following 35\% to `medium density' (32) and the highest 15\% to `high density' (64). Note that even high-density patches are allocated many fewer primitives than number of pixels (196). We learn density-specific convolutional heads $h_{det}$ and $h_{desc}$ to decode detection and description features at varying levels of detail.

\paragraph{From 2D Points to 3D Gaussians.}
The decoder predicts 3D Gaussians in the camera coordinate system of each image. From detected 2D pixel coordinates, we extract depth, RGB colors and descriptors by bilinear interpolation of respectively depth maps, source images and description features. The 3D Gaussian centers $m$ are obtained by unprojection of 2D points using interpolated depth and estimated intrinsics. We simply assign the interpolated color to the primitive, as we observe that it performs similarly to predicting it.  Finally, remaining parameters are predicted by a small MLP from the interpolated descriptors. We decode: 
\begin{itemize}
    \item \textbf{scaling parameters} $s \in \mathbb{R}^{3}$. Instead of direct regression, we define min and max values and interpolate between them using a sigmoid activation. We then multiply scaling parameters by the depth of each Gaussian. 
    This way, the network predicts scale relative to its projected size, rather than an absolute 3D world scale. This ensures that primitives with the same 2D footprint—such as an object $n$ times larger and $n$ times farther away—are represented by a consistent, depth-independent scale parameter.
    \item \textbf{orientations} $q \in \mathbb{R}^{4}$, parametrized as quaternions.
    \item \textbf{opacity} $\alpha \in [0,1]$, obtained from a sigmoid activation.
    \item \textbf{confidence} $c \in [0,1]$, obtained from a sigmoid activation and used during multi-view fusion to select primitives.
\end{itemize}

\begin{figure}[t]
\centering
\includegraphics[width=0.49\textwidth]{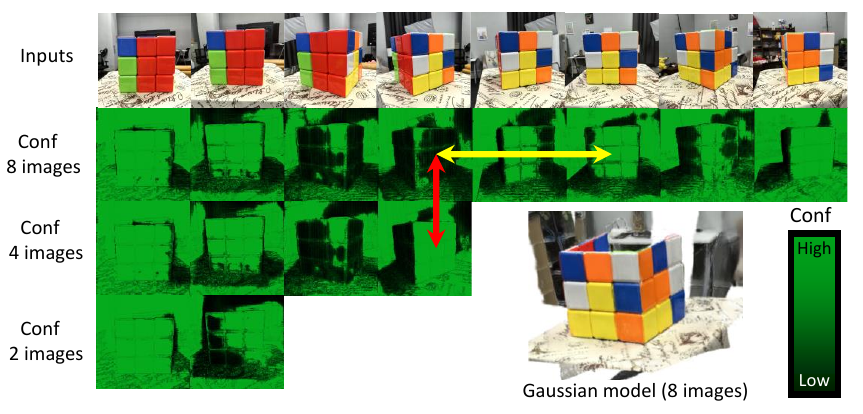}
\caption{\textbf{Confidence maps depending on number of views.} Our model shows multi-view awareness when predicting confidence, removing primitives which are better observed in other views. In the example, one face of the cube is viewed from the side in image 4 and from the front in image 6. When the model sees image 6, Gaussians from image 4 are discarded. Green is high confidence.}
\label{fig:conf}
\end{figure}

\paragraph{Multi-view Aggregation.} 
We build the final 3D Gaussian model by transforming the Gaussian centers $m$ and orientations $q$ from camera to world using predicted extrinsics parameters. Naively gathering primitives from all images leads to redundant Gaussians representing the same content multiple times, leading to blurriness. To address this, we multiply opacities $\alpha$ by confidences $c$, giving the model the ability to prune primitives that would deteriorate the model. We observe, as shown in Figure~\ref{fig:conf}, that the model implicitly learns multi-view reasoning by setting low confidence values to primitives that are observed better in other images. At test-time, we prune primitives for which $\alpha c <0.1$, enabling further computational efficiency.

\paragraph{Image Rendering.} We use a customized 3DGS rasterizer able to render depth and normals (direction of primitives shortest axis), similar to RaDeGS~\cite{zhang2024rade}, that also supports backward pass for camera pose parameters. During training, we render each image $i$ two times, once using only primitives predicted from $i$ (``self-rendering"), and another with the full model. Self-rendering helps to guide the detection process and also enables the model to learn confidence.

\subsection{Training Procedure}
Our method is trained from images only without any 3D annotation. Each training step processes a batch of scenes with a varying number of input images (2 to 12). We describe the loss function we use below. More details on training implementation are given in supplementary materials.

\paragraph{Photometric Losses.}
Our model is trained to render photorealistic images with common photometric supervision losses L1, SSIM~\cite{wang2004image} and LPIPS~\cite{zhang2018unreasonable}. Notably, our system does not require held-out target views as we only render input images. This design is usually avoided because it can lead to a collapsed geometry, but we observe that our teacher geometry losses prevent this problem.

\begin{figure}[t]
\centering
\includegraphics[width=\columnwidth]{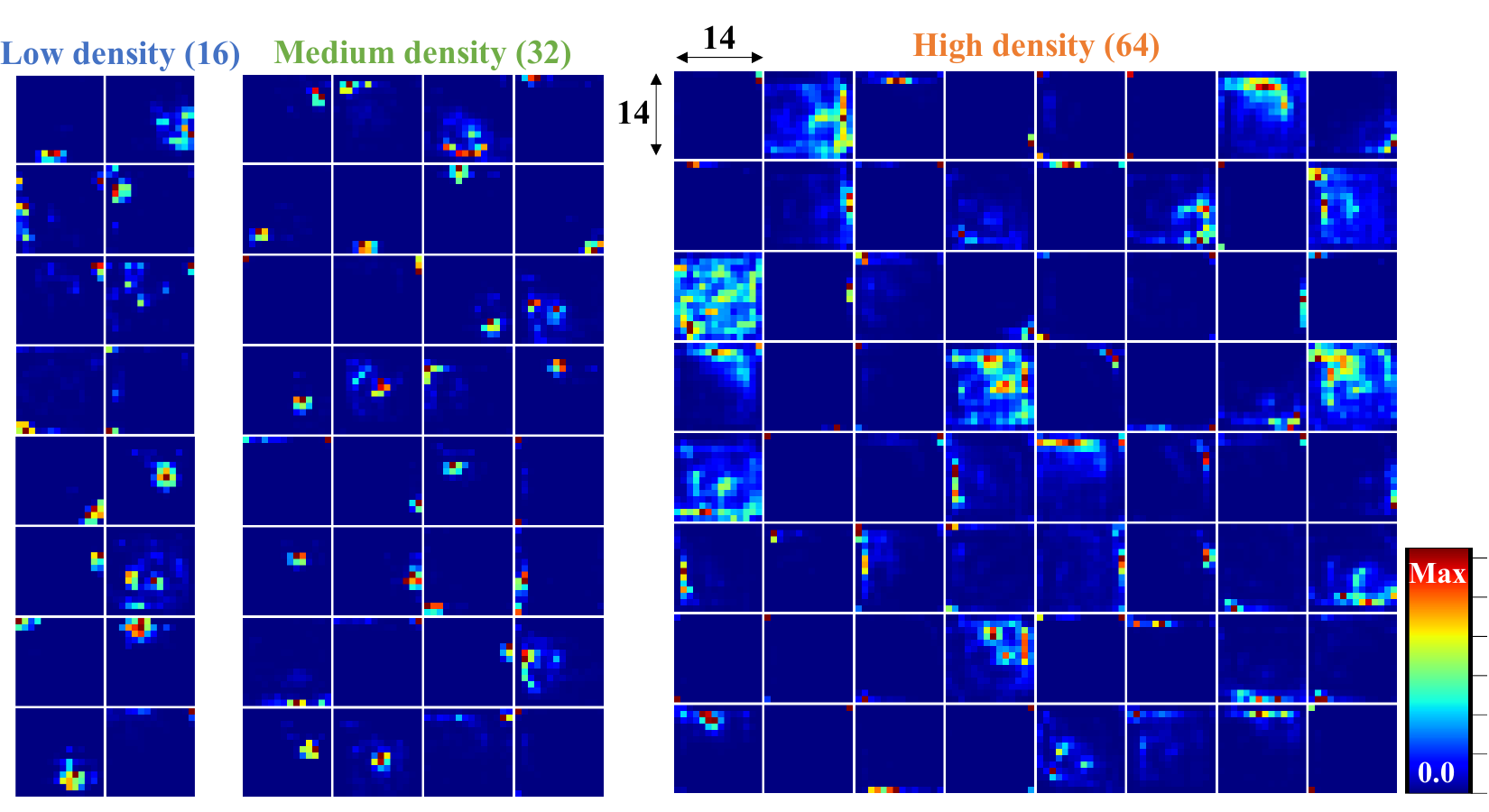}
\label{fig:detection_visu}
\caption{\textbf{Spatial distribution of detection across image patches.} We observe the distribution of our heatmaps $H$ that are used to compute the detected Gaussians. For each density level, we display the average activation of each channel. Most Gaussians appear to operate on a local area of the patch, especially at low density. At high density, some channels are specialized for borders or corners, some others have a widespread distribution enabling to be allocated dynamically to highly detailed areas.}
\label{fig:detection}
\end{figure}

\paragraph{Geometry Consistency Losses.} 
We enforce our 3D Gaussian model and the 3D reconstruction backbone to be consistent with each other. We first define a L1 loss between the predicted and rendered depth maps $L_{depth}$. Then, we apply a second-order constraint $L_{normal}$ by deriving normal maps from the predicted depth maps and intrinsics~\cite{zhang2024rade}. This normal map is compared to rendered normals, defined as the direction of the Gaussians' shortest axis. This encourages Gaussian orientations to align with local surface geometry. 
These consistency losses supervise the Gaussian model geometry, penalizing primitives that reproject wrongly in other images and provides  supervision signal for the depth head, pushing it towards multi-view consistent depth estimation, necessary to obtain accurate Gaussian models.

\paragraph{Teacher Geometry Losses.} Similar to AnySplat~\cite{jiang2025anysplat}, we observe that fine-tuning solely from photometric loss is not stable and causes the model to diverge. To regularize it, we use VGGT~\cite{wang2025vggt} as a teacher model and constrain depth maps and camera poses to remain close to VGGT geometry. We define a loss on depth maps $L_{teach_{depth}}$, weighted by VGGT depth confidence. Then, we use a L1 loss on camera translation $L_{teach_{t}}$ and minimize the geodesic distance between camera orientations $L_{teach_{R}}$. The objective here is not to \textit{distill} information because we start from the same model, but rather to \textit{regularize} the problem to avoid diverging to collapsed geometries.

\begin{figure*}[t]
\centering
 \includegraphics[width=0.98\textwidth]{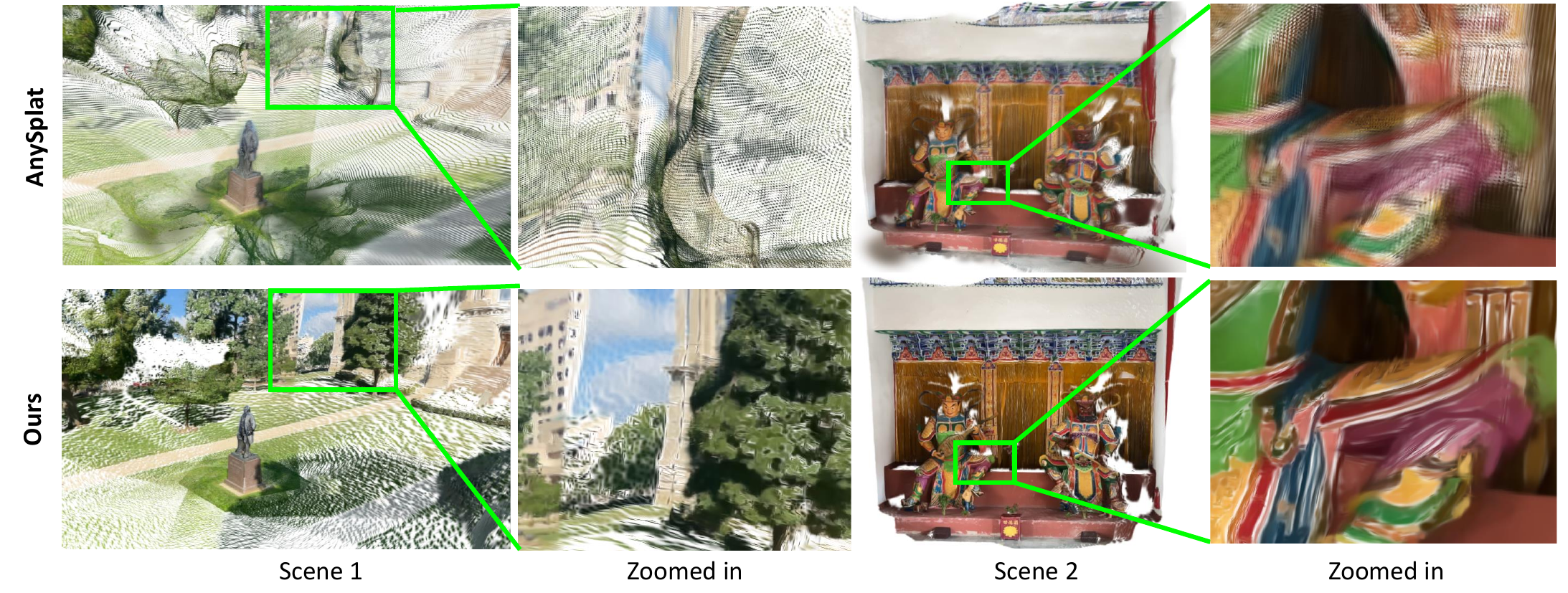}
\caption{\textbf{Extrapolated novel views.} We compare AnySplat and our Gaussian models under highly extrapolated views. AnySplat shows blurriness and oversmoothed geometry while our model shows clean geometries and renderings. When zoomed in, voxel-aligned Gaussians appear visibly and fail to fit the details, in contrast with our detected Gaussians. Models created using 12 views from DL3DV-benchmark.}
\label{fig:extrap_anysplat}
\end{figure*}

\paragraph{Regularization Losses.} We observed opaque objects to be often predicted half-transparent by our model due to the learning of confidence. Consequently, we regularize opacities towards either 0 or 1, using the following loss: $L_{op} = \sum_{i \in G} \sin(\alpha(i) \cdot c(i))$. Finally, because we train with video data, all images from the same scene share the same camera intrinsics. We observe VGGT to produce slightly inconsistent parameters, so we impose consistency as a soft constraint with the L2 distance to the average intrinsic (we minimize variance of intrinsics across the scene).

\begin{table*}[t]
\caption{Novel view synthesis evaluation of pose-free 3DGS methods. Results averaged over 3, 6, 9 and 12 views. \colorbox{red!23}{Best}, \colorbox{orange!26}{runner-up}.}
\label{tab:posefree_nvs_comparison}
\footnotesize
\centering
\resizebox{\linewidth}{!}{%
\begin{tabular}{l c c *{7}{c c c}}
\toprule 
\multicolumn{1}{c}{} & \multicolumn{1}{c}{} & \multicolumn{1}{c}{} & \multicolumn{3}{c}{Average} & \multicolumn{3}{c}{7scenes} & \multicolumn{3}{c}{Charge} & \multicolumn{3}{c}{DL3DV} & \multicolumn{3}{c}{SCRREAM~\cite{jung2024scrream}} & \multicolumn{3}{c}{TanksandTemples} & \multicolumn{3}{c}{mipnerf360} \\
\cmidrule(lr){4-6} \cmidrule(lr){7-9} \cmidrule(lr){10-12} \cmidrule(lr){13-15} \cmidrule(lr){16-18} \cmidrule(lr){19-21} \cmidrule(lr){22-24}
Models & Decoder & \#G / pix & PSNR\,\uparrowaligned & SSIM\,\uparrowaligned & LPIPS\,\downarrowaligned & PSNR\,\uparrowaligned & SSIM\,\uparrowaligned & LPIPS\,\downarrowaligned & PSNR\,\uparrowaligned & SSIM\,\uparrowaligned & LPIPS\,\downarrowaligned & PSNR\,\uparrowaligned & SSIM\,\uparrowaligned & LPIPS\,\downarrowaligned & PSNR\,\uparrowaligned & SSIM\,\uparrowaligned & LPIPS\,\downarrowaligned & PSNR\,\uparrowaligned & SSIM\,\uparrowaligned & LPIPS\,\downarrowaligned & PSNR\,\uparrowaligned & SSIM\,\uparrowaligned & LPIPS\,\downarrowaligned \\
\cmidrule(lr){1-24}
Ours & Ours & \cellbest $0.1431$ & \cellbest $21.21$ & \cellbest $0.6470$ & \cellbest $0.3532$ & \cellbest $25.13$ & \cellbest $0.8119$ & \cellbest $0.2629$ & \cellbest $21.93$ & \cellbest $0.6928$ & \cellbest $0.3997$ & \cellbest $20.48$ & \cellbest $0.6489$ & \cellbest $0.3163$ & \cellbest $22.00$ & \cellbest $0.7171$ & \cellbest $0.3550$ & \cellbest $19.37$ & \cellbest $0.5982$ & \cellbest $0.3348$ & \cellbest $18.36$ & \cellbest $0.4131$ & $0.4503$ \\
AnySplat~\cite{jiang2025anysplat} & Voxel & \cellsecond $0.8141$ & $17.71$ & $0.5075$ & $0.3937$ & \cellsecond $21.39$ & $0.6906$ & \cellsecond $0.3165$ & $18.28$ & $0.6117$ & \cellsecond $0.4320$ & $17.31$ & $0.4710$ & $0.3611$ & $17.56$ & $0.5510$ & $0.4142$ & $16.28$ & $0.4177$ & $0.3851$ & $15.46$ & $0.3028$ & \cellsecond $0.4529$ \\
DA3 GS~\cite{lin2025depth} & Pixel & $1$ & \cellsecond $18.83$ & \cellsecond $0.5428$ & \cellsecond $0.3834$ & $19.66$ & \cellsecond $0.7141$ & $0.3267$ & \cellsecond $20.35$ & \cellsecond $0.6386$ & $0.4893$ & \cellsecond $18.46$ & \cellsecond $0.5243$ & \cellsecond $0.3209$ & \cellsecond $19.65$ & \cellsecond $0.6156$ & \cellsecond $0.3656$ & \cellsecond $16.94$ & \cellsecond $0.4365$ & \cellsecond $0.3780$ & \cellbest $17.95$ & \cellsecond $0.3278$ & \cellbest $0.4203$ \\
\bottomrule
\end{tabular}
}
\end{table*}

\section{Experiments}
\label{sec:experiments}

\paragraph{Training.} We train our method on a single GPU with 140\,GB of VRAM. Similar to VGGT, each training iteration processes a maximum of 24 images and uses a varying number of images per scene, ranging from 2 to 12. We train the model using monocular video sequences and simply sample frames linearly with a random step size ranging from 5 to 10. We train on subsets of 10 datasets: DL3DV~\cite{ling2024dl3dv}, Co3D-v2~\cite{reizenstein2021common}, WildRGBD~\cite{xia2024rgbd}, BlendedMVS~\cite{yao2020blendedmvs}, UnrealStereo4K~\cite{tosi2021smd}, Real Estate 10k, ARKitScenes~\cite{baruch2021arkitscenes}, DTU~\cite{aanaes2016large} and ScanNet++~\cite{Dai_2017_CVPR} and KITTI360~\cite{Liao2022PAMI}. Because no annotation is needed, the process of integrating more training datasets is facilitated. Most of this data is also used for training VGGT~\cite{wang2025vggt} and AnySplat~\cite{jiang2025anysplat}, which is the most closely related method to ours.

\begin{figure*}[h]
    \centering
    \raisebox{-.5\height}{\makebox[0.01\textwidth]{\hspace{-0.2cm} \rotatebox{90}{}}}\enspace
    \makebox[\kubric]{\scriptsize AnySplat~\cite{jiang2025anysplat}}\hfill
    \makebox[\kubric]{\scriptsize DA3 Giant~\cite{lin2025depth}}\hfill
    \makebox[\kubric]{\scriptsize Ours}\hfill
    \makebox[\kubric]{\scriptsize Ground-Truth}\hfill
    \vspace{0.05cm}

    \raisebox{-.5\height}{\makebox[0.01\textwidth]{\hspace{-0.2cm} \rotatebox{90}{\scriptsize 9 views}}}\enspace
    \raisebox{-.5\height}{\includegraphics[width=\kubric]{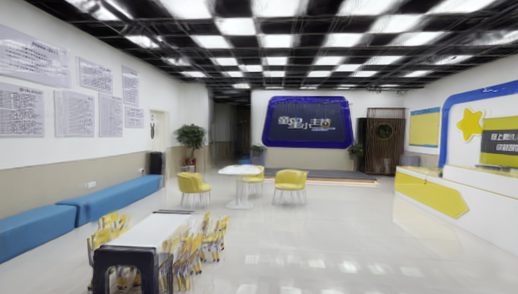}}\hfill
    \raisebox{-.5\height}{\includegraphics[width=\kubric]{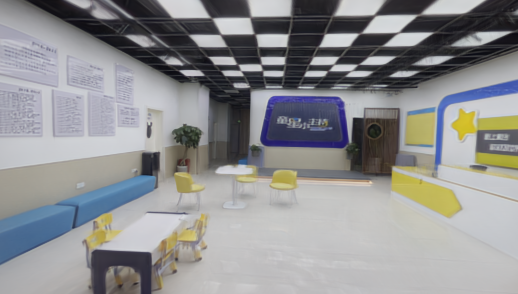}}\hfill
    \raisebox{-.5\height}{\includegraphics[width=\kubric]{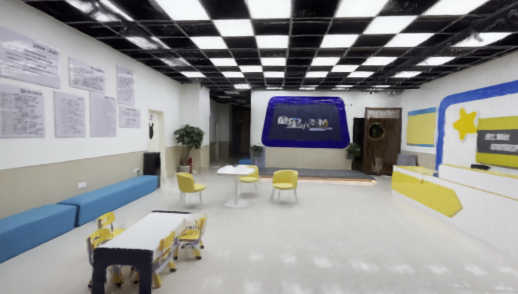}}\hfill
    \raisebox{-.5\height}{\includegraphics[width=\kubric]{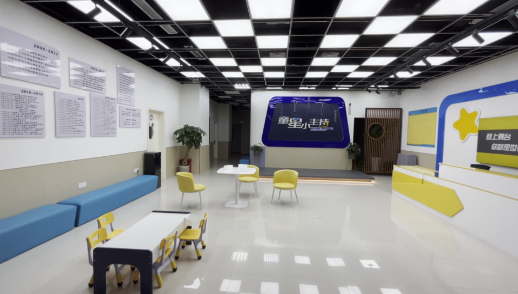}}\hfill
    \vspace{0.05cm}
    
    \raisebox{-.5\height}{\makebox[0.01\textwidth]{\hspace{-0.2cm} \rotatebox{90}{\scriptsize 9 views}}}\enspace
    \raisebox{-.5\height}{\includegraphics[width=\kubric]{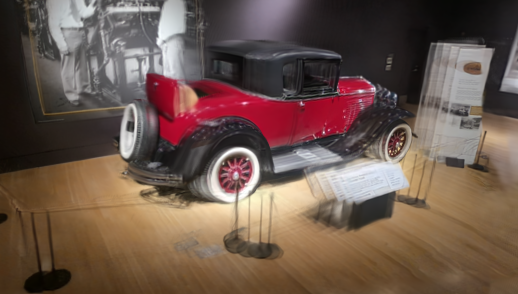}}\hfill
    \raisebox{-.5\height}{\includegraphics[width=\kubric]{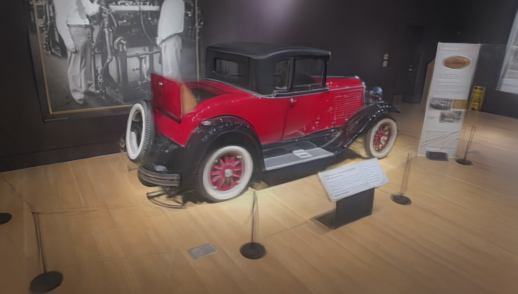}}\hfill
    \raisebox{-.5\height}{\includegraphics[width=\kubric]{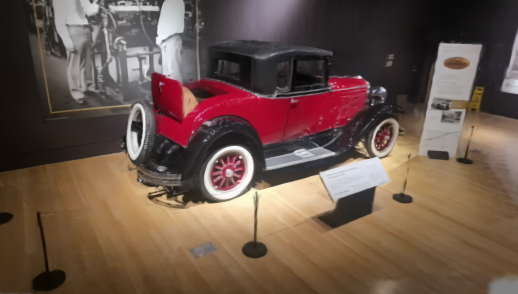}}\hfill
    \raisebox{-.5\height}{\includegraphics[width=\kubric]{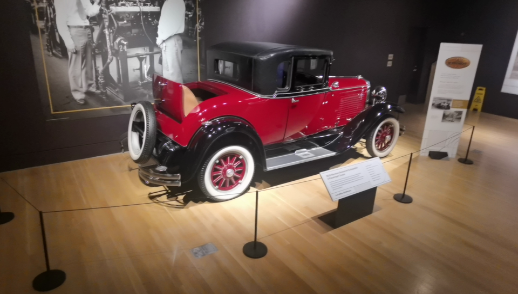}}\hfill
    \vspace{0.05cm}
    \raisebox{-.5\height}{\makebox[0.01\textwidth]{\hspace{-0.2cm} \rotatebox{90}{\scriptsize 3 views}}}\enspace
    \raisebox{-.5\height}{\includegraphics[width=\kubric]{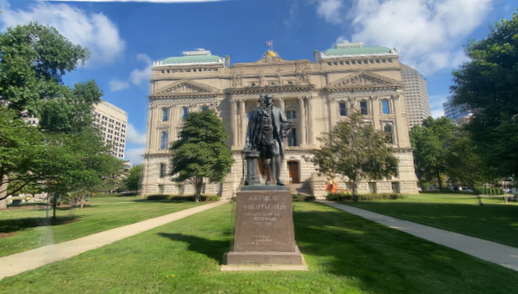}}\hfill
    \raisebox{-.5\height}{\includegraphics[width=\kubric]{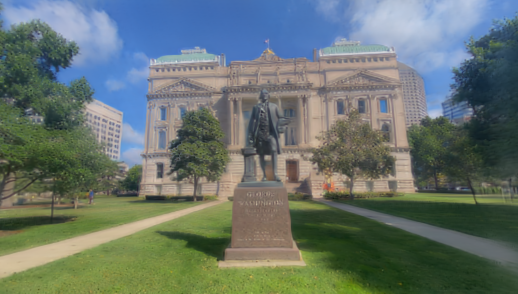}}\hfill
    \raisebox{-.5\height}{\includegraphics[width=\kubric]{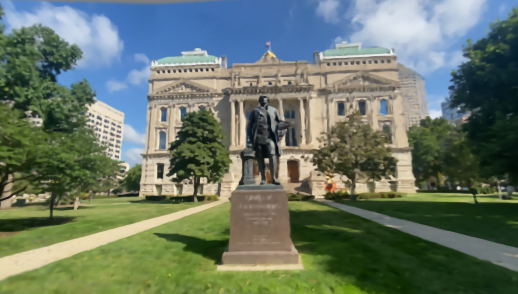}}\hfill
    \raisebox{-.5\height}{\includegraphics[width=\kubric]{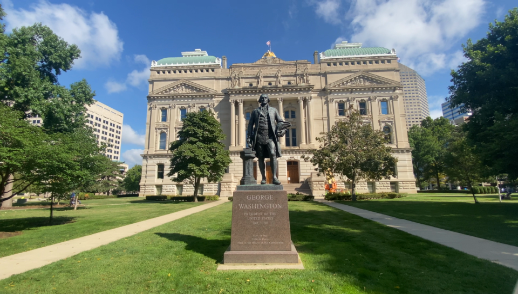}}\hfill
    \vspace{0.05cm}
        \raisebox{-.5\height}{\makebox[0.01\textwidth]{\hspace{-0.2cm} \rotatebox{90}{\scriptsize 3 views}}}\enspace
    \raisebox{-.5\height}{\includegraphics[width=\kubric]{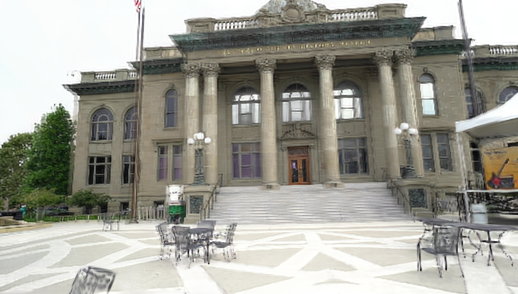}}\hfill
    \raisebox{-.5\height}{\includegraphics[width=\kubric]{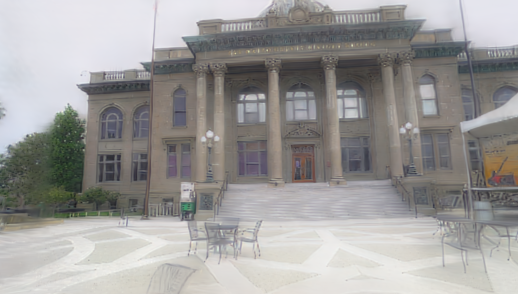}}\hfill
    \raisebox{-.5\height}{\includegraphics[width=\kubric]{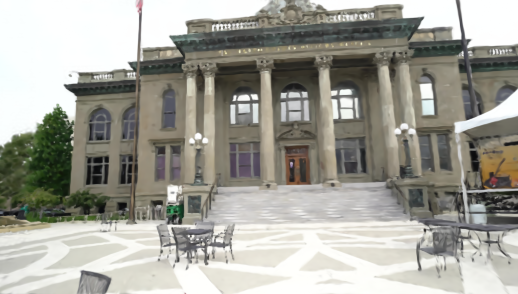}}\hfill
    \raisebox{-.5\height}{\includegraphics[width=\kubric]{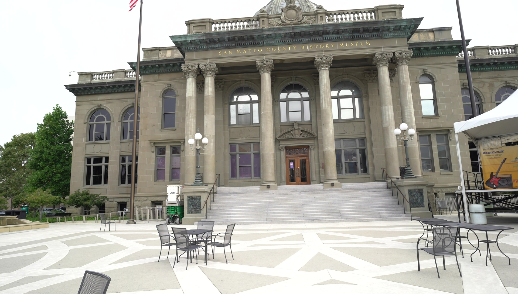}}\hfill
    \vspace{0.05cm}
    \raisebox{-.5\height}{\makebox[0.01\textwidth]{\hspace{-0.2cm} \rotatebox{90}{\scriptsize 6 views}}}\enspace
    \raisebox{-.5\height}{\includegraphics[width=\kubric]{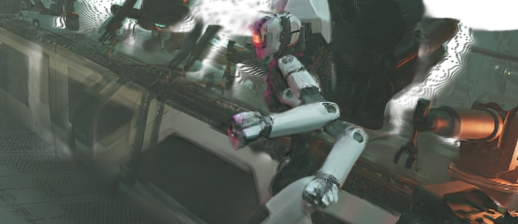}}\hfill
    \raisebox{-.5\height}{\includegraphics[width=\kubric]{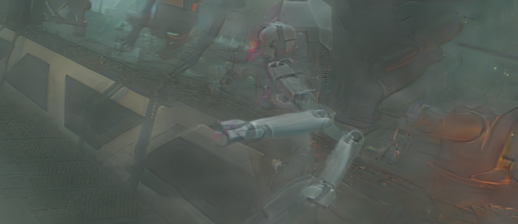}}\hfill
    \raisebox{-.5\height}{\includegraphics[width=\kubric]{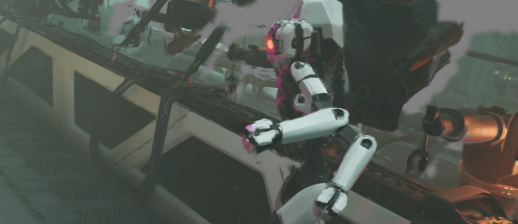}}\hfill
    \raisebox{-.5\height}{\includegraphics[width=\kubric]{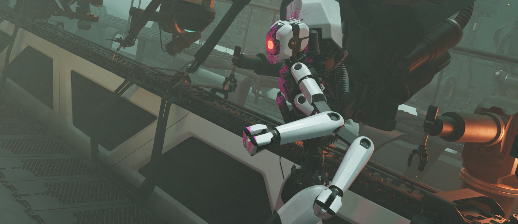}}\hfill
    \vspace{0.05cm}
    \raisebox{-.5\height}{\makebox[0.01\textwidth]{\hspace{-0.2cm} \rotatebox{90}{\scriptsize 25 views}}}\enspace
    \raisebox{-.5\height}{\includegraphics[width=\kubric]{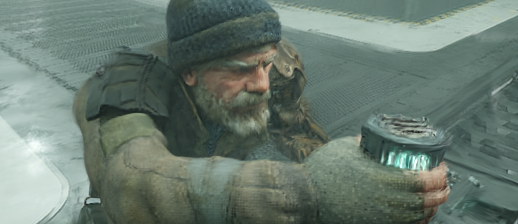}}\hfill
    \raisebox{-.5\height}{\includegraphics[width=\kubric]{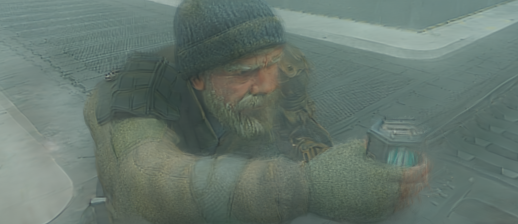}}\hfill
    \raisebox{-.5\height}{\includegraphics[width=\kubric]{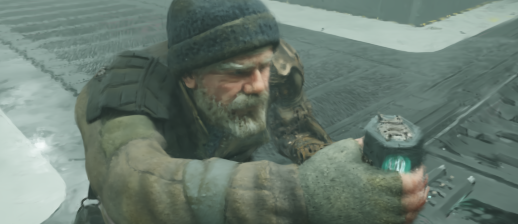}}\hfill
    \raisebox{-.5\height}{\includegraphics[width=\kubric]{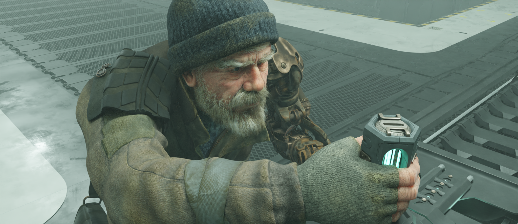}}\hfill
    \vspace{0.05cm}
    
    \caption{\textbf{Novel view synthesis of Pose-Free 3DGS methods.} Zoom in for details.}
    \label{fig:nvs}
    \vspace{-20pt}
\end{figure*}

\subsection{Comparison with Pose-free Methods}

We compare with the most recent feed-forward approaches for 3DGS. AnySplat~\cite{jiang2025anysplat} is the closest work to ours that also fine-tunes VGGT to predict Gaussians. The main difference is the decoder, where they predict voxel-aligned Gaussians. DepthAnything3~\cite{lin2025depth} is another large reconstruction model more recent than VGGT that also include a pixel-aligned 3D Gaussians decoder. We use DA3-GIANT-1.1, a model with 1.15B parameters (1B for VGGT) as a baseline.

\paragraph{Evaluation Datasets.} We use 6 held-out datasets for evaluation, representing a wide variety of captures and imagery.  DL3DV benchmark~\cite{ling2024dl3dv} contains 140 indoor and outdoor scenes captured from handheld cameras.  Charge~\cite{nazarczuk2026charge} is a high-fidelity synthetic dataset rendered from Blender movies which includes ground-truth depth maps and camera poses. SCRREAM~\cite{} 7Scenes~\cite{shotton2013scene},Tanks and Temples~\cite{Knapitsch2017} and MipNeRF360~\cite{barron2022mip} are common NVS benchmarks. Our evaluation uses 3, 6, 9, and 12 views uniformly sampled from the video sequence, with source and target views interleaved. With the exception of Charge, for which we use the predefined sparse (3, 6, 9) and dense (25) splits. We resize images such that their highest dimension is 518 pixels.

\paragraph{Test-time Alignment.} Evaluating pose-free methods for novel view synthesis is challenging because each method predicts the Gaussian model and camera poses in its own coordinate system. To render a target view for evaluation, test-time alignment needs to be performed with reference poses. Aligning with ground-truth poses with standard Umeyama alignment~\cite{umeyama2002least} is often imprecise in sparse settings because of noise in the predicted poses. We employ the same strategy as AnySplat: Given N context views and T target views, we first process our Gaussian model using context views. Then a second independent forward pass is performed using N+T views, to obtain target camera poses. Because VGGT uses the first image as reference, the two models are aligned up to a single scaling factor (if predictions are consistent). This scale is computed by comparing context poses and used to adjusts the predicted target pose. In contrast with prior practice, we do not perform test-time optimization of camera parameters through rendering but evaluate the feed-forward performance.

\paragraph{Novel view synthesis.} Rendering results are evaluated quantitavely in Table~\ref{tab:posefree_nvs_comparison}, our method significantly outperforms AnySplat and DepthAnything3 on all datasets. In Figure~\ref{fig:nvs}, we show qualitative results. DepthAything3 present detailed gaussian models but suffer from bluriness and half-transparency. This common problem is addressed by our opacity regularization loss. AnySplat suffer from geometric inaccuracy, leading to camera misalignment (camera pose) and duplicate geometry (intrinsics) in some scenes. Our approach represents the scene faithfully, even if some details with very thin geometry can disappear. Beyond NVS improvements on interpolated views, the models produced by our method appear `cleaner' under extrapolated viewpoints, avoiding scan-line artefacts commonly observed with pixel-aligned primitives. A visualization is shown in Fig~\ref{fig:extrap_anysplat}. We provide more visualizations in the \href{https://www.youtube.com/watch?v=BlW-wFmHzD4}{supplementary video}.

\paragraph{Compression ratio} In Table~\ref{tab:posefree_nvs_comparison}, we report the ratio \small{$\#G / pix$} between number of primitives and number of input pixels. Pixel-aligned approaches present the densest models with a ratio of 1 but are not more accurate. AnySplat performs voxel-based aggregation to reduce the number of primitives, but the decrease is limited to 19\%. Our models achieves a decrease of 86\% after confidence-based pruning, which correspond to 7 times less primitives than pixel-aligned approaches. Note that this ratio can be controlled by changing parameters of the Adaptive Density mechanism. Our compression reduces the number of primitives per image, and could potentially be combined with views aggregation methods~\cite{wang2025zpressor}.

\paragraph{Geometry evaluation} We evaluate the geometric accuracy of encoders in Table~\ref{tab:geometry_evaluation} using Charge~\cite{nazarczuk2026charge} and SCRREAM~\cite{jung2024scrream} datasets. We report average relative depth error \small{$AbsRel$}, \small{$AUC@30$} for camera pose estimation, and field of view angular error \small{$ang_err$} for intrinsics evaluation. OffTheGrid and AnySplat both use VGGT as encoder, such that reported scores measure the impact of fine-tuning with photometric loss, that enforces multi-view consistent geometry. We observe that, on average, our method successfully improves VGGT for both depth and camera parameters estimation. In contrast, AnySplat fine-tuning degrades the encoder. DA3-Giant is the best encoder for depth and camera pose estimation but the focal length estimation is inaccurate, which explain the bluriness observed in Figure~\ref{fig:nvs}. Our fine-tuned encoder performs the best for intrinsics estimation, allowing to unproject primitives to 3D space accurately.


\begin{table}[t]
\caption{Geometry evaluation on Charge and SCRREAM datasets.}
\label{tab:geometry_evaluation}
\footnotesize
\centering
\resizebox{\linewidth}{!}{%
\begin{tabular}{l *{3}{c c c}}
\toprule 
\multicolumn{1}{c}{} & \multicolumn{3}{c}{Charge} & \multicolumn{3}{c}{SCRREAM} & \multicolumn{3}{c}{Average} \\
\cmidrule(lr){2-4} \cmidrule(lr){5-7} \cmidrule(lr){8-10}
& depth & cam pose & FoV & depth & cam pose & FoV & depth & cam pose & FoV \\
Models & AbsRel\,\downarrowaligned & AUC@30\,\uparrowaligned & ang\_err\,\downarrowaligned & AbsRel\,\downarrowaligned & AUC@30\,\uparrowaligned & ang\_err\,\downarrowaligned & AbsRel\,\downarrowaligned & AUC@30\,\uparrowaligned & ang\_err\,\downarrowaligned \\
\cmidrule(lr){1-10}
OffTheGrid & \cellsecond $0.1609$ & \cellbest $0.9363$ & \cellbest $1.47$ & $0.1258$ & \cellsecond $0.9194$ & \cellbest $0.44$ & \cellsecond $0.1433$ & \cellsecond $0.9278$ & \cellbest $0.96$ \\
DA3-Giant & \cellbest $0.1545$ & \cellsecond $0.9177$ & $6.73$ & \cellbest $0.1133$ & \cellbest $0.95$ & \cellsecond $0.59$ & \cellbest $0.1339$ & \cellbest $0.9339$ & $3.66$ \\
AnySplat & $0.1767$ & $0.8821$ & $2.22$ & $0.1414$ & $0.7845$ & $4.23$ & $0.159$ & $0.8333$ & $3.225$ \\
VGGT & $0.178$ & $0.9012$ & \cellsecond $1.61$ & \cellsecond $0.1201$ & $0.8903$ & $1.37$ & $0.14905$ & $0.8957$ & \cellsecond $1.49$ \\
\bottomrule
\end{tabular}
}
\end{table}

\subsection{Ablation Study}

\begin{figure}[t]
    \centering
    \raisebox{-.5\height}{\makebox[0.01\textwidth]{\hspace{-0.2cm} \rotatebox{90}{}}}\enspace
    \makebox[\kubricr]{\scriptsize AnySplat}\hfill
    \makebox[\kubricr]{\scriptsize Pixel-Aligned}\hfill
    \makebox[\kubricr]{\scriptsize Pixel+Offset }\hfill
    \makebox[\kubricr]{\scriptsize Ours }\hfill
    \makebox[\kubricr]{\scriptsize Ground-Truth}
    \vspace{0.05cm}

    \raisebox{-.5\height}{\makebox[0.01\textwidth]{\hspace{-0.2cm} \rotatebox{90}{\scriptsize 3 views}}}\enspace
    \raisebox{-.5\height}{\includegraphics[width=\kubricr,height=\kubricr]{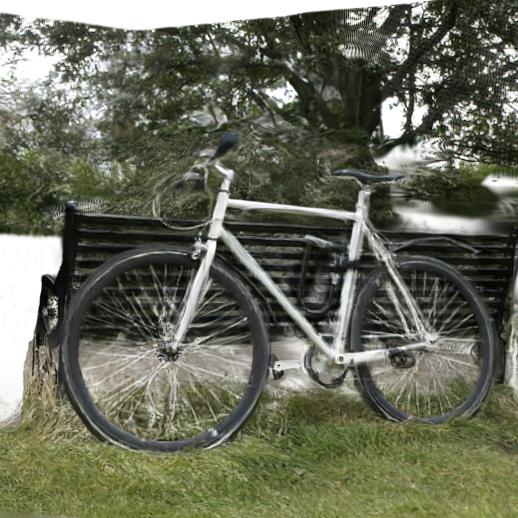}}\hfill
    \raisebox{-.5\height}{\includegraphics[width=\kubricr,height=\kubricr]{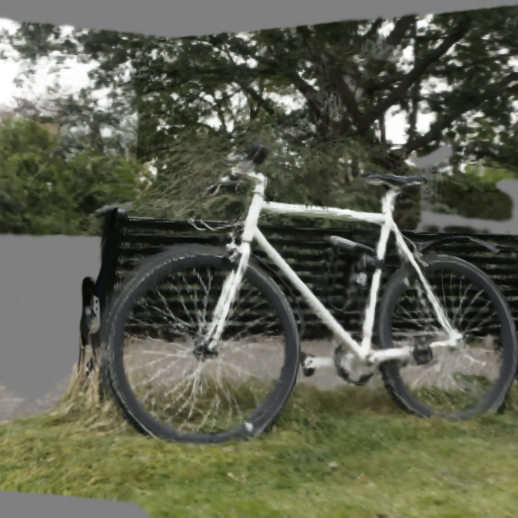}}\hfill
    \raisebox{-.5\height}{\includegraphics[width=\kubricr,height=\kubricr]{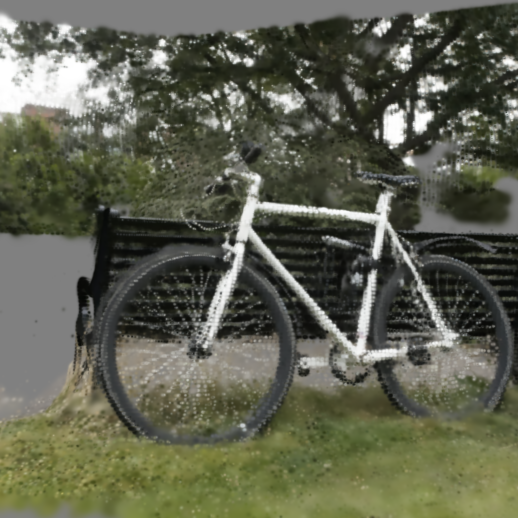}}\hfill
    \raisebox{-.5\height}{\includegraphics[width=\kubricr,height=\kubricr]{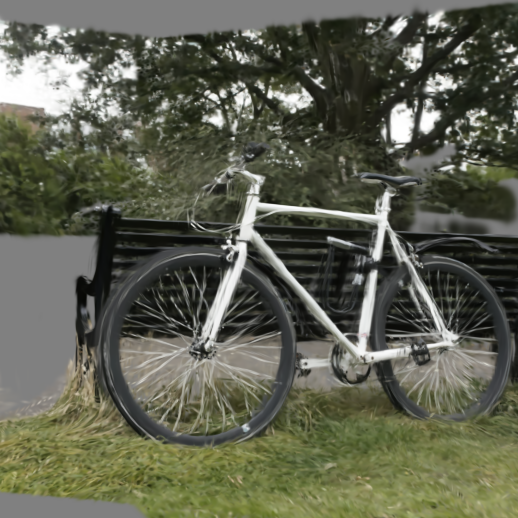}}\hfill
    \raisebox{-.5\height}{\includegraphics[width=\kubricr,height=\kubricr]{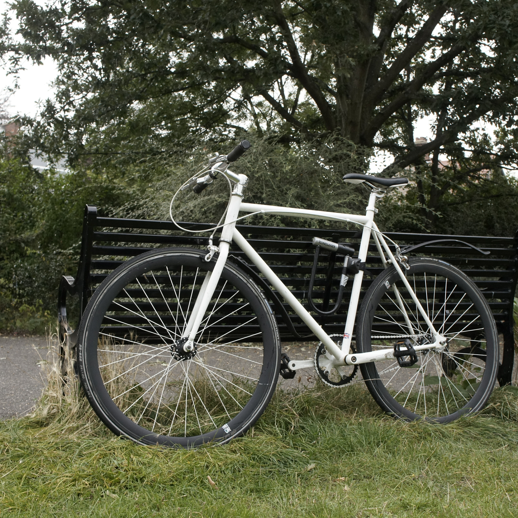}}\hfill
    \raisebox{-.5\height}
    
    \raisebox{-.5\height}{\makebox[0.01\textwidth]{\hspace{-0.2cm} \rotatebox{90}{\scriptsize 6 views}}}\enspace
    \raisebox{-.5\height}{\includegraphics[width=\kubricr,height=\kubricr]{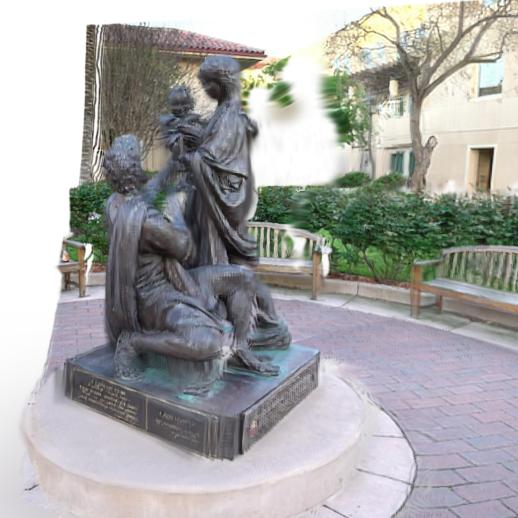}}\hfill
    \raisebox{-.5\height}{\includegraphics[width=\kubricr,height=\kubricr]{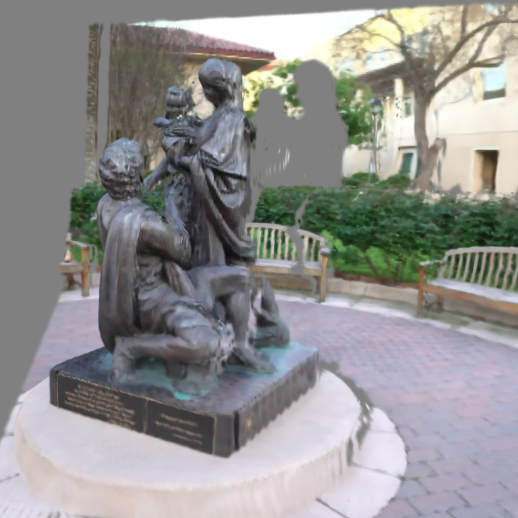}}\hfill
    \raisebox{-.5\height}{\includegraphics[width=\kubricr,height=\kubricr]{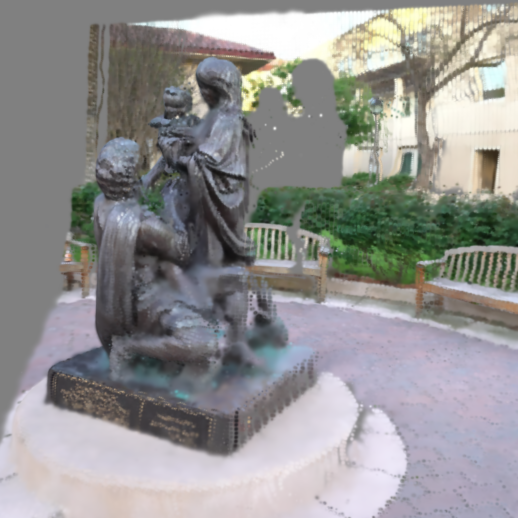}}\hfill
    \raisebox{-.5\height}{\includegraphics[width=\kubricr,height=\kubricr]{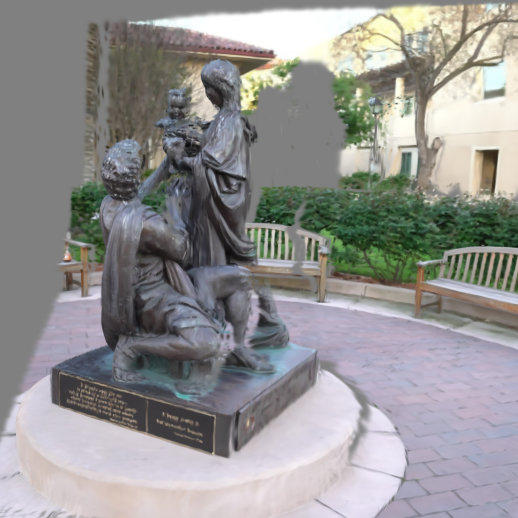}}\hfill
    \raisebox{-.5\height}{\includegraphics[width=\kubricr,height=\kubricr]{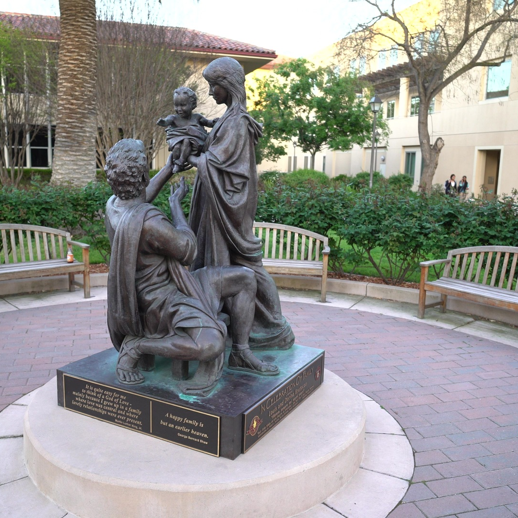}}\hfill

    \raisebox{-.5\height}{\makebox[0.01\textwidth]{\hspace{-0.2cm} \rotatebox{90}{\scriptsize 9 views}}}\enspace
    \raisebox{-.5\height}{\includegraphics[width=\kubricr,height=\kubricr]{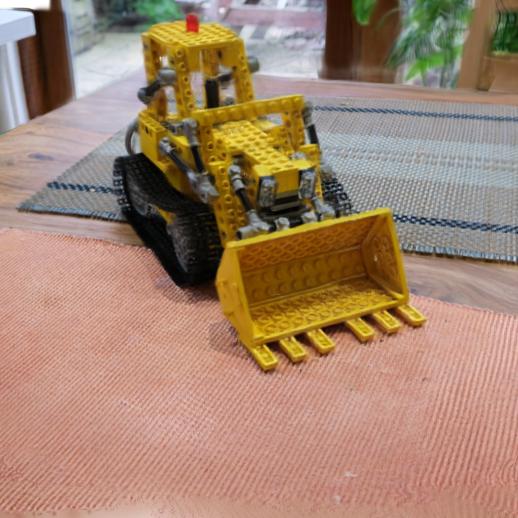}}\hfill
    \raisebox{-.5\height}{\includegraphics[width=\kubricr,height=\kubricr]{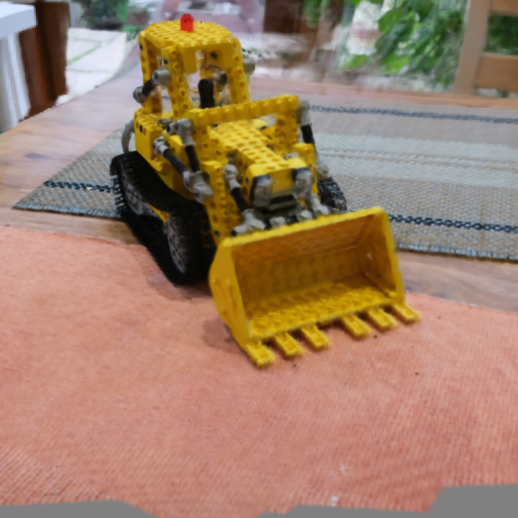}}\hfill
    \raisebox{-.5\height}{\includegraphics[width=\kubricr,height=\kubricr]{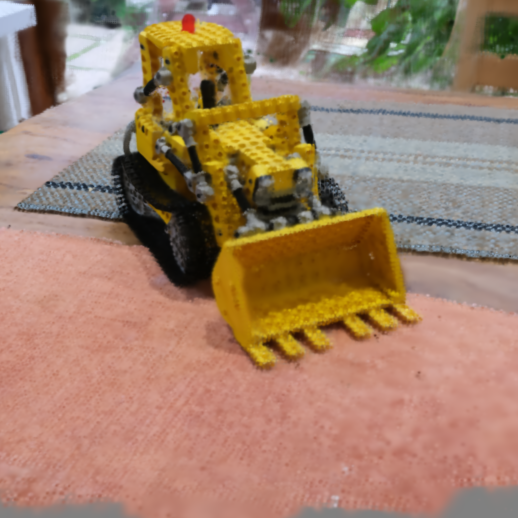}}\hfill
    \raisebox{-.5\height}{\includegraphics[width=\kubricr,height=\kubricr]{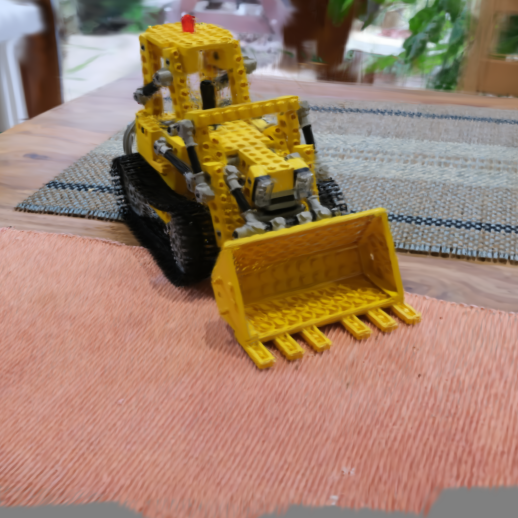}}\hfill
    \raisebox{-.5\height}{\includegraphics[width=\kubricr,height=\kubricr]{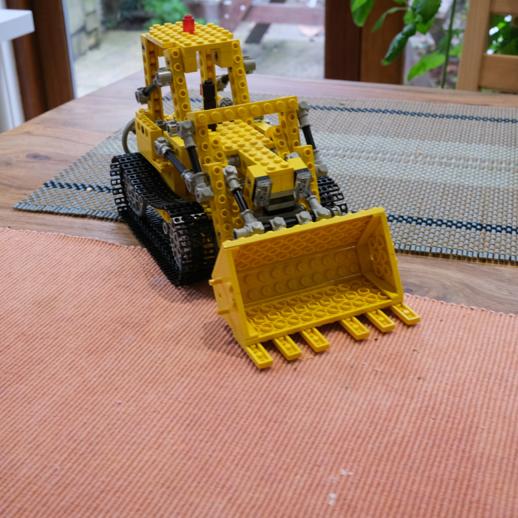}}\hfill

    \raisebox{-.5\height}{\makebox[0.01\textwidth]{\hspace{-0.2cm} \rotatebox{90}{\scriptsize 12 views}}}\enspace
    \raisebox{-.5\height}{\includegraphics[width=\kubricr,height=\kubricr]{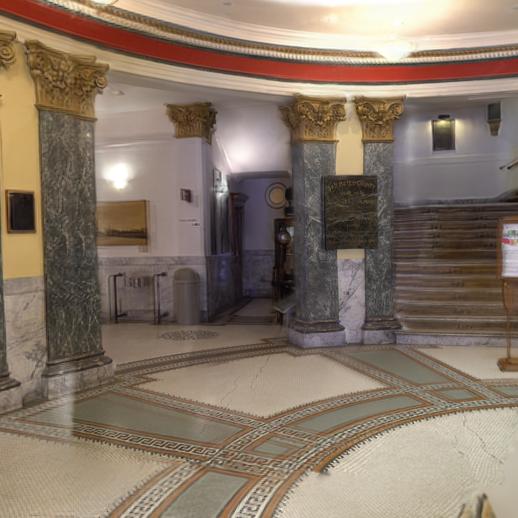}}\hfill
    \raisebox{-.5\height}{\includegraphics[width=\kubricr,height=\kubricr]{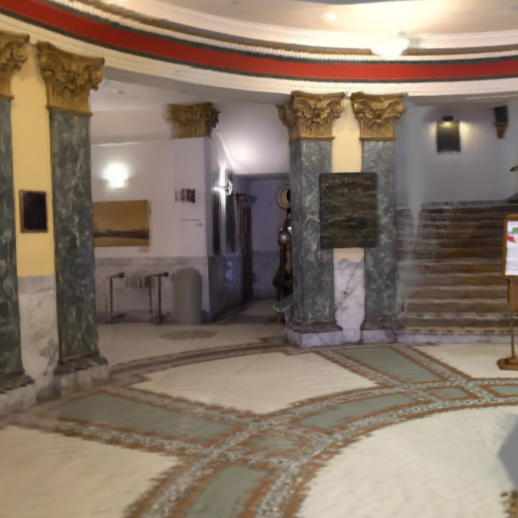}}\hfill
    \raisebox{-.5\height}{\includegraphics[width=\kubricr,height=\kubricr]{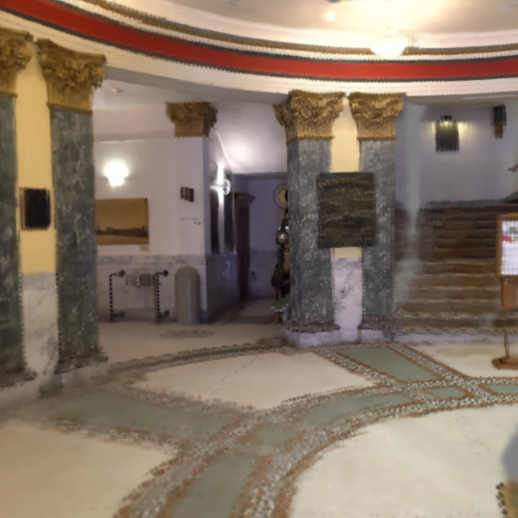}}\hfill
    \raisebox{-.5\height}{\includegraphics[width=\kubricr,height=\kubricr]{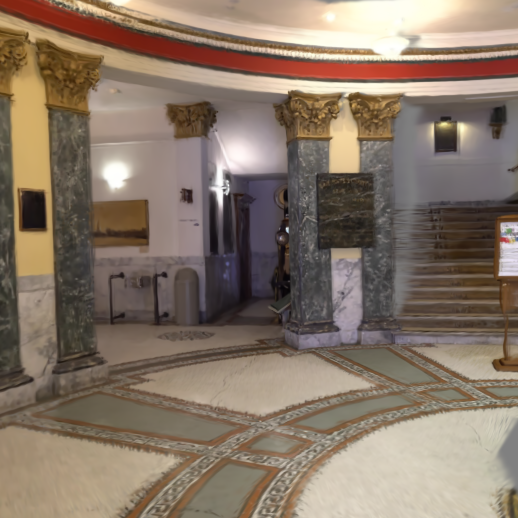}}\hfill
    \raisebox{-.5\height}{\includegraphics[width=\kubricr,height=\kubricr]{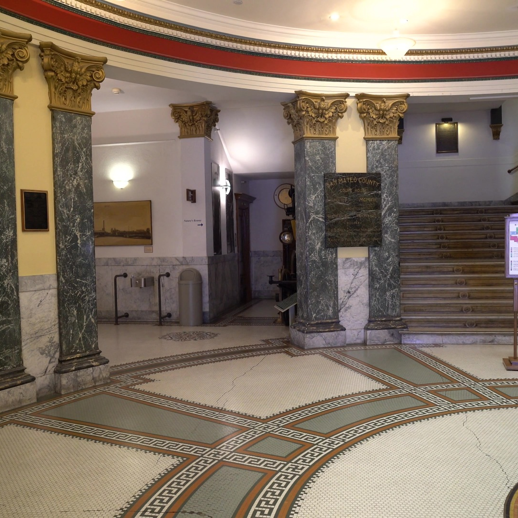}}\hfill
    
    \caption{\textbf{Qualitative Ablation Study.} Zoom in for details.}
    \label{fig:qualitative-results-rebuttal}
    \vspace{-2pt}
\end{figure}

\paragraph{Primitives placement} We study the impact of our primitive detection technique, \ie the computation of 2D Gaussians positions from the DSNT operation, against a pixel-aligned baseline that compute parameters for each pixel from the UNet output (see Fig~\ref{fig:decoder}). One other existing design, proposed by SplatterImage~\cite{szymanowicz2024splatter_image}, is to predict a 3D offset applied to pixel-aligned Gaussians. This way, primitives are also allowed to be positioned anywhere in space. We also implement this design to compare it against ours. Results are reported in Table~\ref{tab:ablation_placement} and can be visualized in Figure~\ref{fig:qualitative-results-rebuttal}. Replacing pixel-aligned by Off-The-Grid provides +0.3dB PSNR, +0.014 SSIM and a 13\% reduction of LPIPS. This difference appears clearly in renderings that appear sharper and less artefacted. The baseline with the offset from SplatterImage presents a similar or degraded rendering quality compared to pixel-aligned, and visible isolated points artefacts, showing that placing primitives accurately requires more advanced techniques.

\begin{table}[h]
    \centering
    \caption{\textbf{Ablation study on primitive placement techniques.}}
    \resizebox{\linewidth}{!}{%
    \begin{tabular}{l ccc ccc ccc}
        \toprule
        & \multicolumn{3}{c}{\textbf{Tanks and Temples}} & \multicolumn{3}{c}{\textbf{MipNeRF360}} & \multicolumn{3}{c}{\textbf{Average}} \\
        \cmidrule(lr){2-4} \cmidrule(lr){5-7} \cmidrule(lr){8-10}
        \textbf{Method} & \textbf{PSNR}$\uparrow$ & \textbf{SSIM}$\uparrow$ & \textbf{LPIPS}$\downarrow$ & \textbf{PSNR}$\uparrow$ & \textbf{SSIM}$\uparrow$ & \textbf{LPIPS}$\downarrow$ & \textbf{PSNR}$\uparrow$ & \textbf{SSIM}$\uparrow$ & \textbf{LPIPS}$\downarrow$ \\
        \midrule
        \textbf{Ours (Full)} & \textbf{19.22} & 0.6044 & \textbf{0.3327} & \textbf{18.03} & \textbf{0.4385} & 0.4265 & \textbf{18.63} & \textbf{0.5215} & \textbf{0.3796} \\
        Ours (No Self Render) & 19.22 & \textbf{0.6074} & 0.3669 & 17.79 & 0.4348 & 0.4614 & 18.51 & 0.5211 & 0.4142 \\
        Pixel Aligned + offset & 18.96 & 0.5810 & 0.4336 & 17.69 & 0.4229 & 0.5151 & 18.33 & 0.5020 & 0.4744 \\
        Pixel Aligned & 19.01 & 0.5877 & 0.3924 & 17.65 & 0.4275 & 0.4757 & 18.33 & 0.5076 & 0.4341 \\
        AnySplat & 16.92 & 0.4919 & 0.3568 & 15.51 & 0.3688 & \textbf{0.4162} & 16.22 & 0.4304 & 0.3865 \\

        \bottomrule
    \end{tabular}%
    }
    
    \label{tab:ablation_placement}
\end{table}

\paragraph{Adaptive density and Confidence-based pruning} We evaluate our model without the adaptive density mechanism (32 gaussians per patch) and without confidence-based pruning on DL3DV benchmark~\cite{ling2024dl3dv}. Novel view synthesis metrics are given in Table~\ref{tab:ablation}. Ablated models perform notably worse, especially in LPIPS. This is because some highly detailed areas can not be fitted accurately with 32 gaussians only. Discarding confidence-based pruning results in duplicated geometry, bluriness and floating artefacts.

\begin{table}[h]
\caption{Ablation on adaptive density and confidence.}
\label{tab:ablation}
\centering
\footnotesize
\setlength{\tabcolsep}{3pt}
\begin{tabularx}{0.7\linewidth}{cc|ZZZ}
\toprule
    {\scriptsize \textbf{Adaptive}} & {\scriptsize \textbf{Confidence}} & \textbf{PSNR} & \textbf{SSIM} & \textbf{LPIPS}  \\ 
    \cmidrule(lr){1-2}\cmidrule(lr){3-5}
        \textbf{\checkmark} & $\times$ & $17.80$ & $0.4223$ & $0.4459$ \\ 
        $\times$ & \textbf{\checkmark} & $18.72$ & $0.5652$ & $0.3898$ \\  
        \textbf{\checkmark} & \textbf{\checkmark} & $\mathbf{19.09}$ & $\mathbf{0.5998}$ & $\mathbf{0.3379}$ \\ 
    \bottomrule
\end{tabularx}
\end{table}


\section{Limitations and Future Work}
\label{sec:discussion}

A limitation of our method (and of most other feed-forward approaches) is that it only reconstructs visible areas, leaving holes in parts of the scene missed during capture and breaking the photorealistic illusion when rendering novel views (for example, the missing top face of the cube in Figure~\ref{fig:conf}). One solution could be to use video diffusion models to produce the final renderings, as proposed by MVSplat360~\cite{chen2024mvsplat360}. Another could be to learn how to complete Gaussian models in a 3D inpainting manner. We leave this problem for future work. Our model currently does not model view-dependent color variations and thus present reduced quality in scenes with lightning variations. Color harmonization could be a solution~\cite{shin2025chroma}. Humans are not represented in the training data and thus our model does not perform well on human subjects, but robustly handles both indoor, outdoor and object centric scenes.
\section{Conclusion}
\label{sec:conclusion}

We have introduced \textit{Off-The-Grid} Gaussians, a novel alternative to pixel-aligned and voxel-aligned strategies that achieves higher photorealism while using far fewer primitives. By combining detection with adaptive density, confidence-based pruning, photometric rendering ang geometry consistency losses, we obtain a model that achieves state-of-the-art accuracy for pose-free feed-forward 3DGS. Our fine-tuning procedure marginally improves the geometry accuracy of VGGT. By decoupling the number of primitives from the number of input pixels, our technique should open the possibility to process 3DGS models from high resolution images with feed-forward methods. 

\clearpage 

{
    \small
    \bibliographystyle{ieeenat_fullname}
    \bibliography{main}

@String{CVPR   = {IEEE Conf. Comput. Vis. Pattern Recog.}}

@String{ECCV   = {Eur. Conf. Comput. Vis.}}

@String{ICCV   = {Int. Conf. Comput. Vis.}}

@String{ICLR   = {Int. Conf. Learn. Represent.}}

@String{NIPS   = {Adv. Neural Inform. Process. Syst.}}

@String{PAMI   = {IEEE Trans. Pattern Anal. Mach. Intell.}}

@String{TOG    = {ACM Trans. Graph.}}

@String(CVPR  = {Computer Vision and Pattern Recognition Conference (CVPR)})

@String(ICCV  = {International Conference on Computer Vision (ICCV)})

@String(ECCV  = {European Conference on Computer Vision (ECCV)})

@String(NIPS  = {Conference on Neural Information Processing Systems})

@String(ICLR  = {International Conference on Learning Representations (ICLR)})

@article{baruch2021arkitscenes,
  title={{ARKitScenes: A Diverse Real-World Dataset For 3D Indoor Scene Understanding Using Mobile RGB-D Data}},
  author={Baruch, Gilad and Chen, Zhuoyuan and Dehghan, Afshin and Dimry, Tal and Feigin, Yuri and Fu, Peter and Gebauer, Thomas and Joffe, Brandon and Kurz, Daniel and Schwartz, Arik and others},
  journal={arXiv preprint arXiv:2111.08897},
  year={2021}
}

@inproceedings{charatan23pixelsplat,
      title={{pixelSplat: 3D Gaussian Splats from Image Pairs for Scalable Generalizable 3D Reconstruction}},
      author={David Charatan and Sizhe Li and Andrea Tagliasacchi and Vincent Sitzmann},
      year={2024},
      booktitle=CVPR,
}

@inproceedings{chen2025dashgaussian,
  title     = {{DashGaussian: Optimizing 3D Gaussian Splatting in 200 Seconds}},
  author    = {Chen, Youyu and Jiang, Junjun and Jiang, Kui and Tang, Xiao and Li, Zhihao and Liu, Xianming and Nie, Yinyu},
  booktitle = CVPR,
  year      = {2025}
}

@inproceedings{chen2024mvsplat,
  title={{MVSplat: Efficient 3D Gaussian Splatting from Sparse Multi-View Images}},
  author={Chen, Yuedong and Xu, Haofei and Zheng, Chuanxia and Zhuang, Bohan and Pollefeys, Marc and Geiger, Andreas and Cham, Tat-Jen and Cai, Jianfei},
  booktitle=ECCV,
  pages={370--386},
  year={2024},
}

@inproceedings{chen2024mvsplat360,
    title     = {{MVSplat360: Feed-Forward 360 Scene Synthesis from Sparse Views}},
    author    = {Chen, Yuedong and Zheng, Chuanxia and Xu, Haofei and Zhuang, Bohan and Vedaldi, Andrea and Cham, Tat-Jen and Cai, Jianfei},
    booktitle = NIPS,
    year      = {2024},
}

@article{choudhury2025accelerating,
  title={{Accelerating Vision Transformers with Adaptive Patch Sizes}},
  author={Choudhury, Rohan and Kim, JungEun and Park, Jinhyung and Yang, Eunho and Jeni, L{\'a}szl{\'o} A and Kitani, Kris M},
  journal={arXiv preprint arXiv:2510.18091},
  year={2025}
}

@InProceedings{Dai_2017_CVPR,
author = {Dai, Angela and Chang, Angel X. and Savva, Manolis and Halber, Maciej and Funkhouser, Thomas and Niessner, Matthias},
title = {{ScanNet: Richly-Annotated 3D Reconstructions of Indoor Scenes}},
booktitle = CVPR,
month = {July},
year = {2017}
}

@article{deng2025,
      title={{Improving Densification in 3D Gaussian Splatting for High-Fidelity Rendering}}, 
      author={Xiaobin Deng and Changyu Diao and Min Li and Ruohan Yu and Duanqing Xu},
      year={2025},
      journal={ arXiv:2508.12313},
}

@inproceedings{detone2018superpoint,
  title={{SuperPoint: Self-Supervised Interest Point Detection and Description}},
  author={DeTone, Daniel and Malisiewicz, Tomasz and Rabinovich, Andrew},
  booktitle=CVPR,
  year={2018}
}

@article{fan2024instantsplat,
  title={{InstantSplat: Sparse-view Gaussian Splatting in Seconds}},
  author={Fan, Zhiwen and Cong, Wenyan and Wen, Kairun and Wang, Kevin and Zhang, Jian and Ding, Xinghao and Xu, Danfei and Ivanovic, Boris and Pavone, Marco and Pavlakos, Georgios and others},
  journal={arXiv preprint arXiv:2403.20309},
  year={2024}
}

@inproceedings{fu2024colmap,
  title={{COLMAP-Free 3D Gaussian Splatting}},
  author={Fu, Yang and Liu, Sifei and Kulkarni, Amey and Kautz, Jan and Efros, Alexei A and Wang, Xiaolong},
  booktitle=CVPR,
  year={2024}
}

@article{hong2024pf3plat,
    title   = {{PF3plat: Pose-Free Feed-Forward 3D Gaussian Splatting}},
    author  = {Sunghwan Hong and Jaewoo Jung and Heeseong Shin and Jisang Han and Jiaolong Yang and Chong Luo and Seungryong Kim},
    journal = {arXiv preprint arXiv:2410.22128},
    year    = {2024}
}

@inproceedings{huang2025no,
  title={{No Pose at All: Self-Supervised Pose-Free 3D Gaussian Splatting from Sparse Views}},
  author={Huang, Ranran and Mikolajczyk, Krystian},
  booktitle=CVPR,
  year={2025}
}

@inproceedings{jiang2024construct,
  title={{A Construct-Optimize Approach to Sparse View Synthesis without Camera Pose}},
  author={Jiang, Kaiwen and Fu, Yang and Varma T, Mukund and Belhe, Yash and Wang, Xiaolong and Su, Hao and Ramamoorthi, Ravi},
  booktitle={ACM SIGGRAPH Conference Papers},
  year={2024}
}

@article{jiang2025anysplat,
  title={{AnySplat: Feed-forward 3D Gaussian Splatting from Unconstrained Views}},
  author={Jiang, Lihan and Mao, Yucheng and Xu, Linning and Lu, Tao and Ren, Kerui and Jin, Yichen and Xu, Xudong and Yu, Mulin and Pang, Jiangmiao and Zhao, Feng and others},
  journal={arXiv preprint arXiv:2505.23716},
  year={2025}
}

@inproceedings{
jin2025lvsm,
title={{LVSM: A Large View Synthesis Model with Minimal 3D Inductive Bias}},
author={Haian Jin and Hanwen Jiang and Hao Tan and Kai Zhang and Sai Bi and Tianyuan Zhang and Fujun Luan and Noah Snavely and Zexiang Xu},
booktitle=ICLR,
year={2025},
}

@inproceedings{kang2025selfsplat,
  title={{SelfSplat: Pose-Free and 3D Prior-Free Generalizable 3D Gaussian Splatting}},
  author={Kang, Gyeongjin and Yoo, Jisang and Park, Jihyeon and Nam, Seungtae and Im, Hyeonsoo and Shin, Sangheon and Kim, Sangpil and Park, Eunbyung},
  booktitle=CVPR,
  year={2025}
}

@inproceedings{keetha2025mapanything,
  title       = {{MapAnything}: Universal Feed-Forward Metric {3D} Reconstruction},
  author      = {Nikhil Keetha and Norman M\"uller and Johannes Sch\"onberger and Lorenzo Porzi and
                 Yuchen Zhang and Tobias Fischer and Arno Knapitsch and Duncan Zauss and
                 Ethan Weber and Nelson Antunes and Jonathon Luiten and Manuel Lopez-Antequera and Samuel Rota Bul\`o and Christian Richardt and Deva Ramanan and Sebastian Scherer and Peter Kontschieder},
  booktitle   = {arXiv:2509.13414},
  year        = {2025}
}

@Article{kerbl3Dgaussians,
      author       = {Kerbl, Bernhard and Kopanas, Georgios and Leimk{\"u}hler, Thomas and Drettakis, George},
      title        = {{3D Gaussian Splatting for Real-Time Radiance Field Rendering}},
      journal      = {ACM Transactions on Graphics},
      number       = {4},
      volume       = {42},
      year         = {2023},
}

@inproceedings{kim2024color,
  title={C{olor-cued Efficient Densification Method for 3D Gaussian Splatting}},
  author={Kim, Sieun and Lee, Kyungjin and Lee, Youngki},
  booktitle=CVPR,
  year={2024}
}

@inproceedings{leroy2024grounding,
  title={{Grounding Image Matching in 3D with MASt3R}},
  author={Leroy, Vincent and Cabon, Yohann and Revaud, J{\'e}r{\^o}me},
  booktitle=ECCV,
  year={2024},
}

@article{li2025vicasplat,
  title={{VicaSplat: A Single Run is All You Need for 3D Gaussian Splatting and Camera Estimation from Unposed Video Frames}},
  author={Li, Zhiqi and Dong, Chengrui and Chen, Yiming and Huang, Zhangchi and Liu, Peidong},
  journal={arXiv preprint arXiv:2503.10286},
  year={2025}
}

@inproceedings{ling2024dl3dv,
  title={{DL3DV-10K: A Large-Scale Scene Dataset for Deep Learning-based 3D Vision}},
  author={Ling, Lu and Sheng, Yichen and Tu, Zhi and Zhao, Wentian and Xin, Cheng and Wan, Kun and Yu, Lantao and Guo, Qianyu and Yu, Zixun and Lu, Yawen and others},
  booktitle = CVPR,
  year={2024}
}

@inproceedings{lyu2025resgs,
  title={{ResGS: Residual Densification of 3D Gaussian for Efficient Detail Recovery}},
  author={Lyu, Yanzhe and Cheng, Kai and Kang, Xin and Chen, Xuejin},
  booktitle=CVPR,
  year={2025}
}

@inproceedings{Mallick2024taming3dgs,
    author = {Mallick, Saswat Subhajyoti and Goel, Rahul and Kerbl, Bernhard and Steinberger, Markus and Carrasco, Francisco Vicente and De La Torre, Fernando},
    title = {Taming {3DGS}: High-Quality Radiance Fields with Limited Resources},
    year = {2024},
    booktitle = {SIGGRAPH Asia Conference Papers},
    }

@article{Matsuki2024,
  title={{G}aussian {S}platting {SLAM}},
  author={Hidenobu Matsuki and Riku Murai and Paul H. J. Kelly and Andrew J. Davison},
  journal=CVPR,
  year={2024}
}

@article{meuleman2025fly,
  title={{On-the-fly Reconstruction for Large-Scale Novel View Synthesis from Unposed Images}},
  author={Meuleman, Andreas and Shah, Ishaan and Lanvin, Alexandre and Kerbl, Bernhard and Drettakis, George},
  journal={ACM Transactions on Graphics (TOG)},
  volume={44},
  number={4},
  pages={1--14},
  year={2025},
  publisher={ACM New York, NY, USA}
}

@article{mildenhall2021nerf,
  title={{NeRF: Representing Scenes as Neural Radiance Fields for View Synthesis}},
  author={Mildenhall, Ben and Srinivasan, Pratul P and Tancik, Matthew and Barron, Jonathan T and Ramamoorthi, Ravi and Ng, Ren},
  journal={Communications of the ACM},
  volume={65},
  number={1},
  pages={99--106},
  year={2021},
  publisher={ACM New York, NY, USA}
}

@article{mueller2022instant,
    author = {Thomas M\"uller and Alex Evans and Christoph Schied and Alexander Keller},
    title = {{Instant Neural Graphics Primitives with a Multiresolution Hash Encoding}},
    journal = {ACM Transactions on Graphics (TOG)},
    issue_date = {July 2022},
    volume = {41},
    number = {4},
    month = jul,
    year = {2022},
    pages = {102:1--102:15},
    articleno = {102},
    numpages = {15},
    publisher = {ACM},
}

@article{nibali2018numerical,
  title={{Numerical Coordinate Regression with Convolutional Neural Networks}},
  author={Nibali, Aiden and He, Zhen and Morgan, Stuart and Prendergast, Luke},
  journal={arXiv preprint arXiv:1801.07372},
  year={2018}
}

@article{
oquab2024dinov2,
title={{{DINO}v2: Learning Robust Visual Features without Supervision}},
author={Maxime Oquab and Timoth{\'e}e Darcet and Th{\'e}o Moutakanni and Huy V. Vo and Marc Szafraniec and Vasil Khalidov and Pierre Fernandez and Daniel Haziza and Francisco Massa and Alaaeldin El-Nouby and Mido Assran and Nicolas Ballas and Wojciech Galuba and Russell Howes and Po-Yao Huang and Shang-Wen Li and Ishan Misra and Michael Rabbat and Vasu Sharma and Gabriel Synnaeve and Hu Xu and Herve Jegou and Julien Mairal and Patrick Labatut and Armand Joulin and Piotr Bojanowski},
journal={Transactions on Machine Learning Research},
year={2024},
}

@inproceedings{reizenstein2021common,
  title={{Common Objects in 3D: Large-Scale Learning and Evaluation of Real-life 3D Category Reconstruction}},
  author={Reizenstein, Jeremy and Shapovalov, Roman and Henzler, Philipp and Sbordone, Luca and Labatut, Patrick and Novotny, David},
  booktitle=CVPR,
  year={2021}
}

@article{ren2025fastgs,
      title={{FastGS: Training 3D Gaussian Splatting in 100 Seconds}}, 
      author={Shiwei Ren and Tianci Wen and Yongchun Fang and Biao Lu},
      year={2025},
      journal={arXiv:2511.04283}
}

@inproceedings{ronneberger2015u,
  title={{U-Net: Convolutional networks for biomedical image segmentation}},
  author={Ronneberger, Olaf and Fischer, Philipp and Brox, Thomas},
  booktitle={International Conference on Medical image computing and computer-assisted intervention},
  year={2015},
}

@inproceedings{rota2024revising,
  title={{Revising Densification in Gaussian Splatting}},
  author={Rota Bul{\`o}, Samuel and Porzi, Lorenzo and Kontschieder, Peter},
  booktitle=ECCV,
  year={2024},
}

@inproceedings{schoenberger2016sfm,
    author={Sch\"{o}nberger, Johannes Lutz and Frahm, Jan-Michael},
    title={{Structure-from-Motion Revisited}},
    booktitle=CVPR,
    year={2016}
    }

@inproceedings{7scenes,
  title={Scene coordinate regression forests for camera relocalization in {RGB-D} images},
  author={Shotton, Jamie and Glocker, Ben and Zach, Christopher and Izadi, Shahram and Criminisi, Antonio and Fitzgibbon, Andrew},
  booktitle=CVPR,
  pages={2930--2937},
  year={2013}
}

@article{smart2024splatt3r,
  title={{Splatt3R: Zero-shot Gaussian Splatting from Uncalibrated Image Pairs}},
  author={Smart, Brandon and Zheng, Chuanxia and Laina, Iro and Prisacariu, Victor Adrian},
  journal={arXiv preprint arXiv:2408.13912},
  year={2024}
}

@article{szymanowicz2024splatter_image,
  author    = {Szymanowicz, Stanislaw and Rupprecht, Christian and Vedaldi, Andrea},
  title     = {{Splatter Image: Ultra-Fast Single-View 3D Reconstruction}},
  journal   = CVPR,
  year      = {2024},
}

@inproceedings{tanay2024global,
  title={{Global Latent Neural Rendering}},
  author={Tanay, Thomas and Maggioni, Matteo},
  booktitle=CVPR,
  year={2024}
}

@inproceedings{tosi2021smd,
  title={{SMD-Nets: Stereo Mixture Density Networks}},
  author={Tosi, Fabio and Liao, Yiyi and Schmitt, Carolin and Geiger, Andreas},
  booktitle=CVPR,
  year={2021}
}

@article{umeyama2002least,
  title={Least-squares estimation of transformation parameters between two point patterns},
  author={Umeyama, Shinji},
  journal={IEEE Transactions on Pattern Analysis and Machine Intelligence},
  volume={13},
  number={4},
  pages={376--380},
  year={1991},
  publisher={IEEE}
}

@inproceedings{wang2025vggt,
  title={{VGGT: Visual Geometry Grounded Transformer}},
  author={Wang, Jianyuan and Chen, Minghao and Karaev, Nikita and Vedaldi, Andrea and Rupprecht, Christian and Novotny, David},
  booktitle=CVPR,
  year={2025}
}

@inproceedings{wang2024dust3r,
  title={{DUSt3R: Geometric 3D Vision Made Easy}},
  author={Wang, Shuzhe and Leroy, Vincent and Cabon, Yohann and Chidlovskii, Boris and Revaud, Jerome},
  booktitle=CVPR,
  year={2024}
}

@article{wang2025volsplat,
  title={{VolSplat: Rethinking Feed-Forward 3D Gaussian Splatting with Voxel-Aligned Prediction}},
  author={Wang, Weijie and Chen, Yeqing and Zhang, Zeyu and Liu, Hengyu and Wang, Haoxiao and Feng, Zhiyuan and Qin, Wenkang and Zhu, Zheng and Chen, Donny Y and Zhuang, Bohan},
  journal={arXiv preprint arXiv:2509.19297},
  year={2025}
}

@inproceedings{wang2024freesplat,
  title={{FreeSplat: Generalizable 3D Gaussian Splatting Towards Free-View Synthesis of Indoor Scenes}},
  author={Wang, Yunsong and Huang, Tianxin and Chen, Hanlin and Lee, Gim Hee},
  booktitle=NIPS,
  year={2024}
}

@article{wang2025freesplat++,
  title={{FreeSplat++: Generalizable 3D Gaussian Splatting for Efficient Indoor Scene Reconstruction}},
  author={Wang, Yunsong and Huang, Tianxin and Chen, Hanlin and Lee, Gim Hee},
  journal={arXiv preprint arXiv:2503.22986},
  year={2025}
}

@article{wang2025pi3,
      title={{$\pi^3$: Permutation-Equivariant Visual Geometry Learning}}, 
      author={Yifan Wang and Jianjun Zhou and Haoyi Zhu and Wenzheng Chang and Yang Zhou and Zizun Li and Junyi Chen and Jiangmiao Pang and Chunhua Shen and Tong He},
      journal={arXiv preprint arXiv:2507.13347},
      year={2025}
}

@article{wang2004image,
  title={Image quality assessment: from error visibility to structural similarity},
  author={Wang, Zhou and Bovik, Alan C and Sheikh, Hamid R and Simoncelli, Eero P},
  journal={IEEE Transactions on Image Processing},
  volume={13},
  number={4},
  pages={600--612},
  year={2004},
}

@inproceedings{xu2024depthsplat,
      title   = {DepthSplat: Connecting Gaussian Splatting and Depth},
      author  = {Xu, Haofei and Peng, Songyou and Wang, Fangjinhua and Blum, Hermann and Barath, Daniel and Geiger, Andreas and Pollefeys, Marc},
      booktitle=CVPR,
      year={2025}
    }

@inproceedings{xu20254dgt,
    title     = {{4DGT: Learning a 4D Gaussian Transformer Using Real-World Monocular Videos}},
    author    = {Xu, Zhen and Li, Zhengqin and Dong, Zhao and Zhou, Xiaowei and Newcombe, Richard and Lv, Zhaoyang},
    booktitle   = NIPS,
    year      = {2025}
}

@article{ye2024no,
  title={{No Pose, No Problem: Surprisingly Simple 3D Gaussian Splats from Sparse Unposed Images}},
  author={Ye, Botao and Liu, Sifei and Xu, Haofei and Li, Xueting and Pollefeys, Marc and Yang, Ming-Hsuan and Peng, Songyou},
  journal={arXiv preprint arXiv:2410.24207},
  year={2024}
}

@article{zhang2024rade,
  title={{RaDe-GS: Rasterizing Depth in Gaussian Splatting}},
  author={Zhang, Baowen and Fang, Chuan and Shrestha, Rakesh and Liang, Yixun and Long, Xiaoxiao and Tan, Ping},
  journal={arXiv preprint arXiv:2406.01467},
  year={2024}
}

@article{gslrm2024,
    author={Zhang, Kai and Bi, Sai and Tan, Hao and Xiangli, Yuanbo and Zhao, Nanxuan 
      and Sunkavalli, Kalyan and Xu, Zexiang},
    title     = {{GS-LRM: Large Reconstruction Model for 3D Gaussian Splatting}},
    journal   = ECCV,
    year      = {2024},
}

@inproceedings{zhang2018unreasonable,
  title={{The Unreasonable Effectiveness of Deep Features as a Perceptual Metric}},
  author={Zhang, Richard and Isola, Phillip and Efros, Alexei A and Shechtman, Eli and Wang, Oliver},
  booktitle=CVPR,
  year={2018}
}

@inproceedings{zhang2025flare,
  title={{FLARE: Feed-forward Geometry, Appearance and Camera Estimation from Uncalibrated Sparse Views}},
  author={Zhang, Shangzhan and Wang, Jianyuan and Xu, Yinghao and Xue, Nan and Rupprecht, Christian and Zhou, Xiaowei and Shen, Yujun and Wetzstein, Gordon},
  booktitle=CVPR,
  year={2025}
}

@inproceedings{moreau2024human,
  title={{Human Gaussian Splatting: Real-time Rendering of Animatable Avatars}},
  author={Moreau, Arthur and Song, Jifei and Dhamo, Helisa and Shaw, Richard and Zhou, Yiren and P{\'e}rez-Pellitero, Eduardo},
  booktitle=CVPR,
  year={2024}
}

@inproceedings{jiang2024hifi4g,
  title={{HiFi4G: High-Fidelity Human Performance Rendering via Compact Gaussian Splatting}},
  author={Jiang, Yuheng and Shen, Zhehao and Wang, Penghao and Su, Zhuo and Hong, Yu and Zhang, Yingliang and Yu, Jingyi and Xu, Lan},
  booktitle=CVPR,
  pages={19734--19745},
  year={2024}
}

@inproceedings{jiang2025rayzer,
   title={RayZer: A Self-supervised Large View Synthesis Model},
   author={Jiang, Hanwen and Tan, Hao and Wang, Peng and Jin, Haian and Zhao, Yue and Bi, Sai and Zhang, Kai and Luan, Fujun and Sunkavalli, Kalyan and Huang, Qixing and Pavlakos, Georgios},
   booktitle=ICCV,
   year={2025},
}

@misc{milesial2025pytorchunet,
  author       = {Mile Sial},
  title        = {{Pytorch-UNet: PyTorch implementation of the U-Net for image semantic segmentation}},
  howpublished = {\url{https://github.com/milesial/Pytorch-UNet/}},
  year         = {2025},
  note         = {Accessed: 2025-11-20}
}

@article{jung2024scrream,
  title={Scrream: Scan, register, render and map: A framework for annotating accurate and dense 3d indoor scenes with a benchmark},
  author={Jung, HyunJun and Li, Weihang and Wu, Shun-Cheng and Bittner, William and Brasch, Nikolas and Song, Jifei and P{\'e}rez-Pellitero, Eduardo and Zhang, Zhensong and Moreau, Arthur and Navab, Nassir and others},
  journal={Advances in Neural Information Processing Systems},
  volume={37},
  pages={44164--44176},
  year={2024}
}

@article{shaw2026ico3d,
      title={ICo3D: An Interactive Conversational 3D Virtual Human},
      author={Shaw, Richard and Jang, Youngkyoon and Papaioannou, Athanasios and Moreau, Arthur and Dhamo, Helisa and Zhang, Zhensong and Pérez-Pellitero, Eduardo},
      journal={International Journal of Computer Vision},
      year={2026},
      volume={134},
      number={4},
      pages={161},
      doi={10.1007/s11263-025-02725-8},
      url={https://doi.org/10.1007/s11263-025-02725-8}
    }

@inproceedings{xia2024rgbd,
  title={Rgbd objects in the wild: Scaling real-world 3d object learning from rgb-d videos},
  author={Xia, Hongchi and Fu, Yang and Liu, Sifei and Wang, Xiaolong},
  booktitle={Proceedings of the IEEE/CVF Conference on Computer Vision and Pattern Recognition},
  pages={22378--22389},
  year={2024}
}

@inproceedings{yao2020blendedmvs,
  title={Blendedmvs: A large-scale dataset for generalized multi-view stereo networks},
  author={Yao, Yao and Luo, Zixin and Li, Shiwei and Zhang, Jingyang and Ren, Yufan and Zhou, Lei and Fang, Tian and Quan, Long},
  booktitle={Proceedings of the IEEE/CVF conference on computer vision and pattern recognition},
  pages={1790--1799},
  year={2020}
}

@article{aanaes2016large,
  title={Large-Scale Data for Multiple-View Stereopsis},
  author={Aan{\ae}s, Henrik and Jensen, Rasmus Ramsb{\o}l and Vogiatzis, George and Tola, Engin and Dahl, Anders Bjorholm},
  journal={International Journal of Computer Vision},
  pages={1--16},
  year={2016},
  publisher={Springer}
}

@inproceedings{nazarczuk2026charge,
      title={Charge: A Comprehensive Novel View Synthesis Benchmark and Dataset to Bind Them All},
      author={Nazarczuk, Michal and Tanay, Thomas and Moreau, Arthur and Zhang, Zhensong and P{\'e}rez-Pellitero, Eduardo},
      booktitle=CVPR,
      year={2026}
    }

@article{Liao2022PAMI,
   title =  {{KITTI}-360: A Novel Dataset and Benchmarks for Urban Scene Understanding in 2D and 3D},
   author = {Yiyi Liao and Jun Xie and Andreas Geiger},
   journal = {Pattern Analysis and Machine Intelligence (PAMI)},
   year = {2022},
}

@inproceedings{barron2022mip,
  title={Mip-nerf 360: Unbounded anti-aliased neural radiance fields},
  author={Barron, Jonathan T and Mildenhall, Ben and Verbin, Dor and Srinivasan, Pratul P and Hedman, Peter},
  booktitle={Proceedings of the IEEE/CVF conference on computer vision and pattern recognition},
  pages={5470--5479},
  year={2022}
}

@article{Knapitsch2017,
    author    = {Arno Knapitsch and Jaesik Park and Qian-Yi Zhou and Vladlen Koltun},
    title     = {Tanks and Temples: Benchmarking Large-Scale Scene Reconstruction},
    journal   = {ACM Transactions on Graphics},
    volume    = {36},
    number    = {4},
    year      = {2017},
}

@inproceedings{shotton2013scene,
  title={Scene coordinate regression forests for camera relocalization in RGB-D images},
  author={Shotton, Jamie and Glocker, Ben and Zach, Christopher and Izadi, Shahram and Criminisi, Antonio and Fitzgibbon, Andrew},
  booktitle={Proceedings of the IEEE conference on computer vision and pattern recognition},
  pages={2930--2937},
  year={2013}
}

@article{lin2025depth,
  title={Depth anything 3: Recovering the visual space from any views},
  author={Lin, Haotong and Chen, Sili and Liew, Junhao and Chen, Donny Y and Li, Zhenyu and Shi, Guang and Feng, Jiashi and Kang, Bingyi},
  journal={arXiv preprint arXiv:2511.10647},
  year={2025}
}

@article{wang2025zpressor,
  title={Zpressor: Bottleneck-aware compression for scalable feed-forward 3dgs},
  author={Wang, Weijie and Chen, Donny Y and Zhang, Zeyu and Shi, Duochao and Liu, Akide and Zhuang, Bohan},
  journal={arXiv preprint arXiv:2505.23734},
  year={2025}
}

@inproceedings{shin2025chroma,
      title={CHROMA: Consistent Harmonization of Multi-View Appearance via Bilateral Grid Prediction},
      author={Shin, Jisu and Shaw, Richard and Shin, Seunghyun and Zhang, Zhensong and Jeon, Hae-Gon and Perez-Pellitero, Eduardo},
      booktitle=ICLR,
      year={2026}
    }
}

\clearpage
\setcounter{page}{1}
\maketitlesupplementary

\section{Supplementary video}

We present novel view synthesis comparison with AnySplat~\cite{jiang2025anysplat} in the \href{https://arthurmoreau.github.io/OffTheGrid/}{supplementary video}. Trajectories include both interpolated and extrapolated views. We show models that use 6 views and generate 30 views interpolated between each. In the middle of the generated video, we extrapolate views with a spiral trajectory. Please note that despite using the same code to generate trajectories for both methods, views are not exactly aligned due to the difference in camera poses. Our method exhibits more accurate geometry and sharper rendering in most scenes. 

\section{Implementation details on our method}

 For the 3D reconstruction backbone, we use VGGT-1B~\cite{wang2025vggt}, starting from official checkpoints released by the authors. We remove the pointmap head that we don't use. We tried to use it instead of depth map but observed significantly inferior accuracy, especially for large number of images. Regarding the decoder, the UNet module uses the implementation of Pytorch-UNet~\cite{milesial2025pytorchunet} with 13 input channels (3 for RGB, 2 for depth and depth confidence, and 8 for unpatchified latent features and 32 output channels. The $h_{det}$ and $h_{desc}$ heads process features with input image resolution through 3 convolutional layers with ReLU intermediate activations and 32 channels. Detection features are transformed into heatmaps through a softmax with temperature $0.2$. Importantly, we remind that softmax is not applied over channels dimension but over spatial dimension at a patch level. Bilinear interpolation is done with Pytorch \texttt{grid\_sample} function with \texttt{padding\_mode} set to border.

\section{Depth estimation}

\begin{figure*}
\setlength{\tabcolsep}{0pt}
\renewcommand{\arraystretch}{0.00}
\begin{centering}
\begin{tabular}{cccccc}
{\scriptsize Input Image} & {\scriptsize Ground Truth} & {\scriptsize VGGT\,\cite{wang2025vggt}} & {\scriptsize AnySplat\,\cite{jiang2025anysplat}} & {\scriptsize DepthSplat\,\cite{xu2024depthsplat}} & {\scriptsize Ours}\\[5pt]
\includegraphics[width=0.166\textwidth]{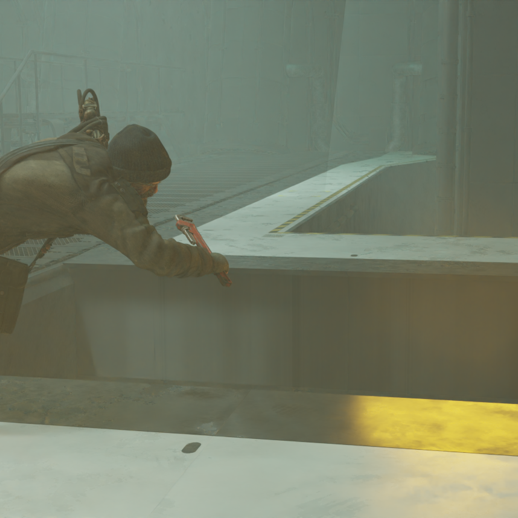} & \includegraphics[width=0.166\textwidth]{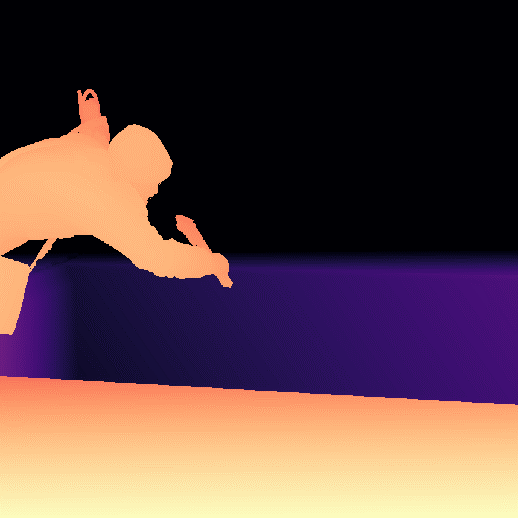} & \includegraphics[width=0.166\textwidth]{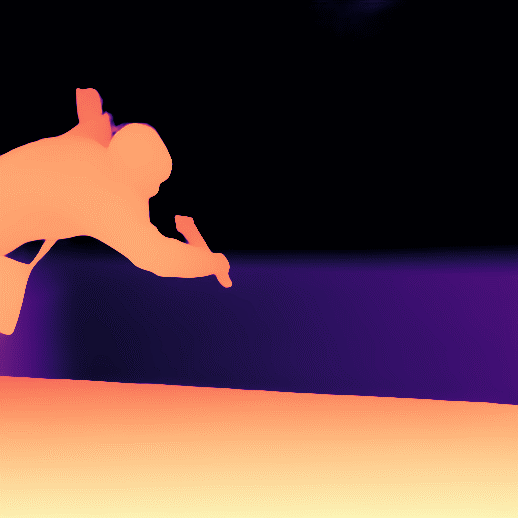} & \includegraphics[width=0.166\textwidth]{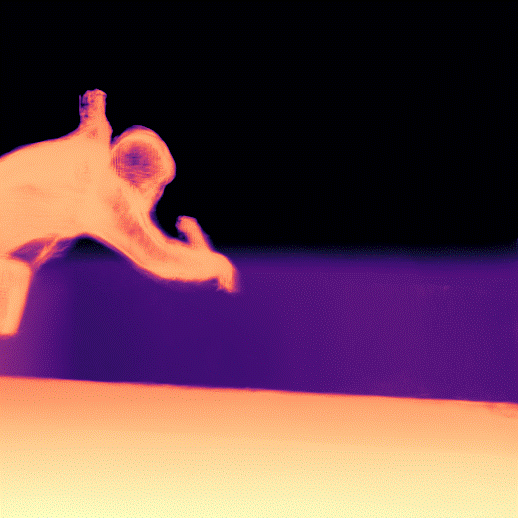} & \includegraphics[width=0.166\textwidth]{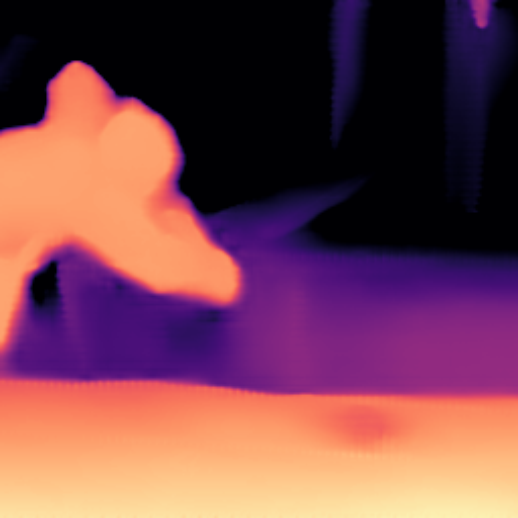} & \includegraphics[width=0.166\textwidth]{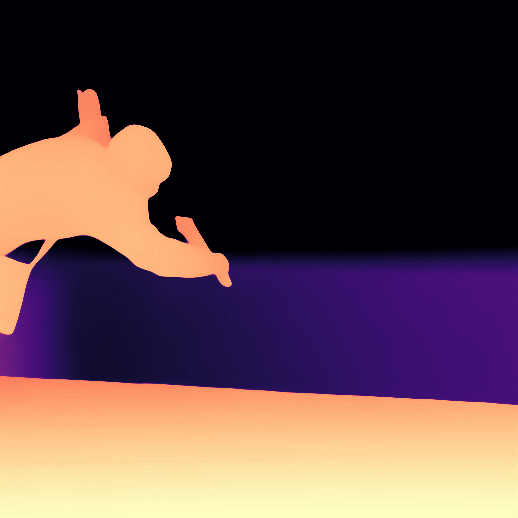}\tabularnewline
\includegraphics[width=0.166\textwidth]{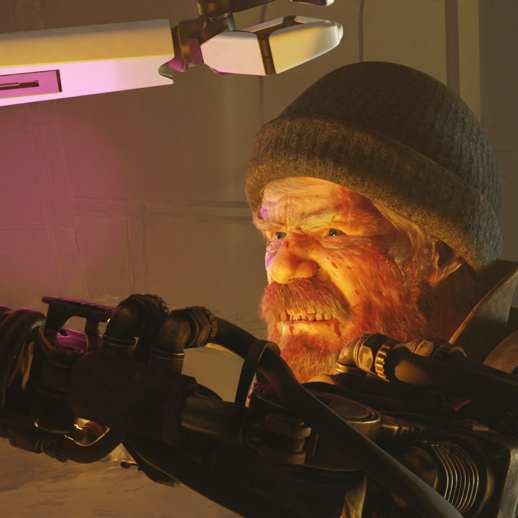} & \includegraphics[width=0.166\textwidth]{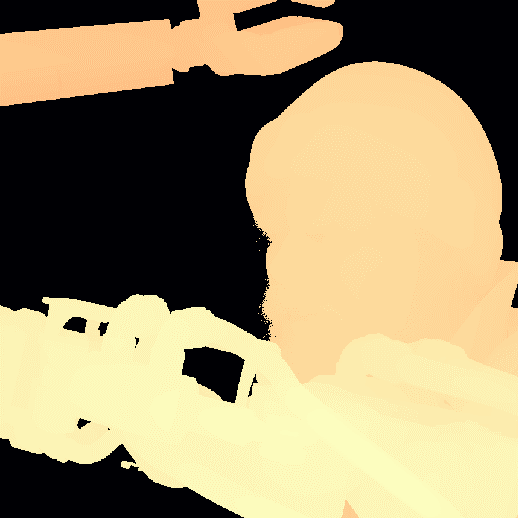} & \includegraphics[width=0.166\textwidth]{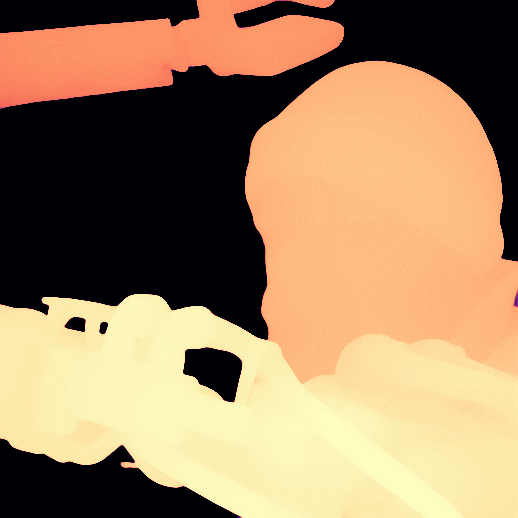} & \includegraphics[width=0.166\textwidth]{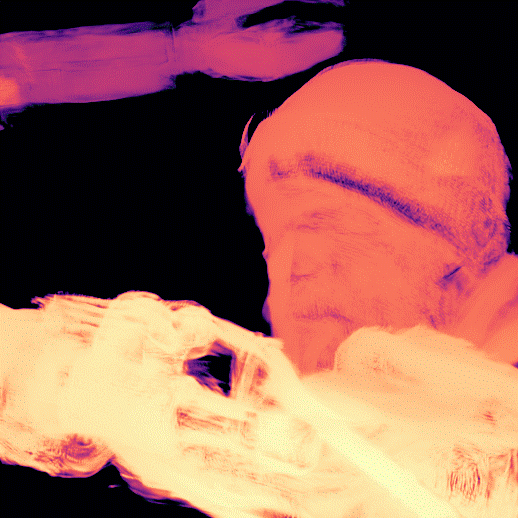} & \includegraphics[width=0.166\textwidth]{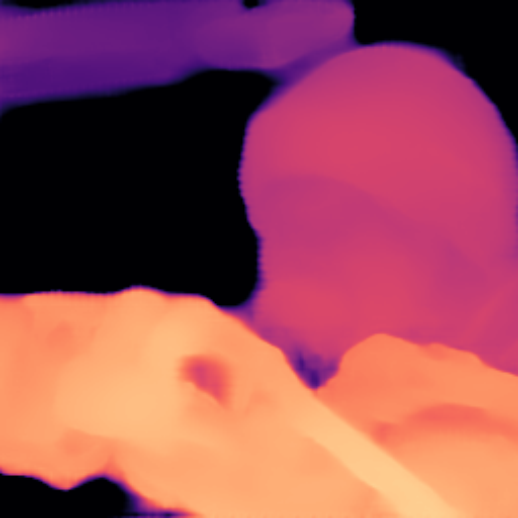} & \includegraphics[width=0.166\textwidth]{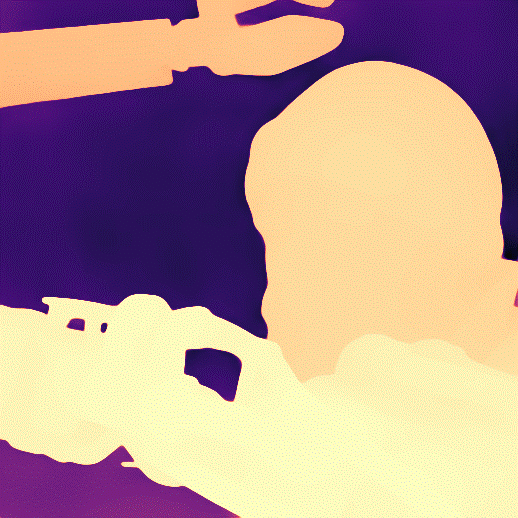}\tabularnewline
\includegraphics[width=0.166\textwidth]{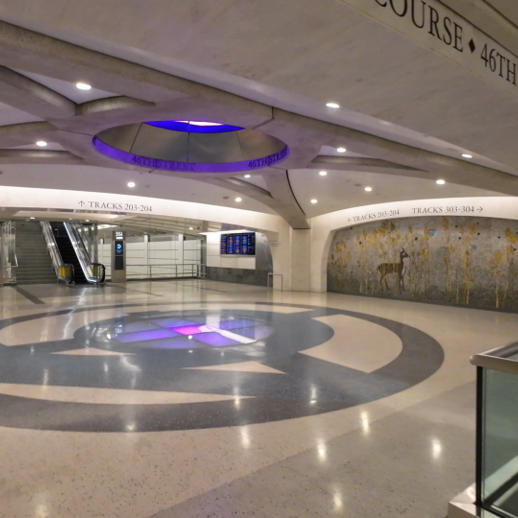} & &
\includegraphics[width=0.166\textwidth]{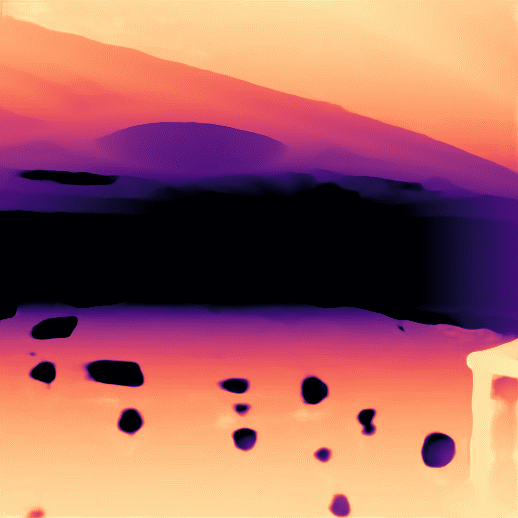} & \includegraphics[width=0.166\textwidth]{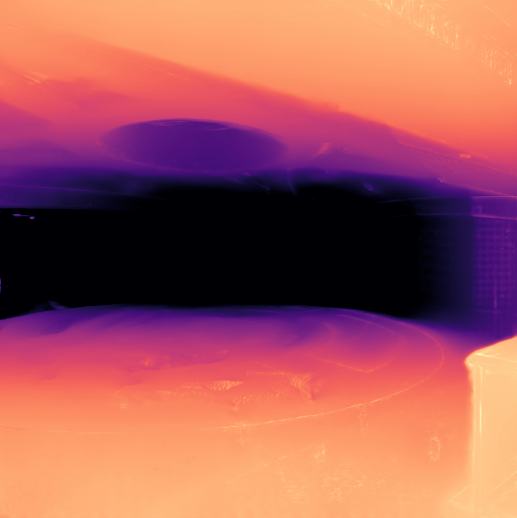} & \includegraphics[width=0.166\textwidth]{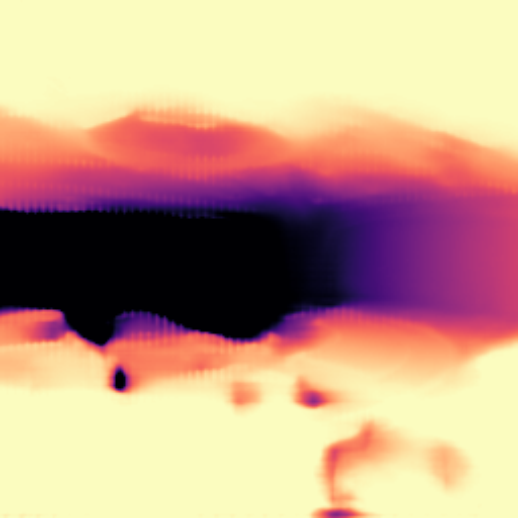} & \includegraphics[width=0.166\textwidth]{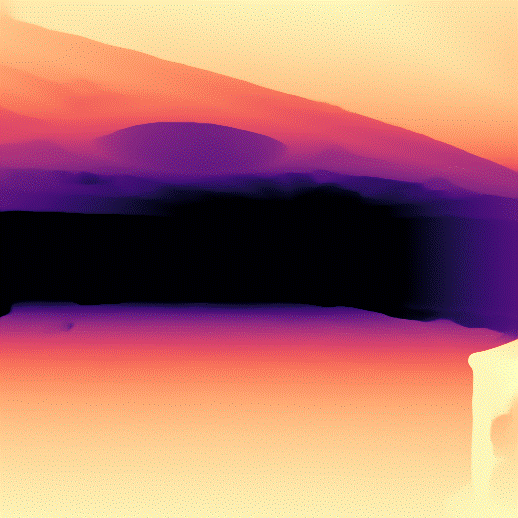}\tabularnewline
\includegraphics[width=0.166\textwidth]{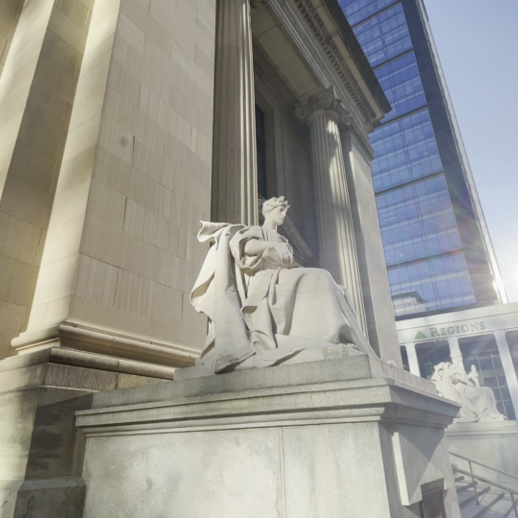} & & \includegraphics[width=0.166\textwidth]{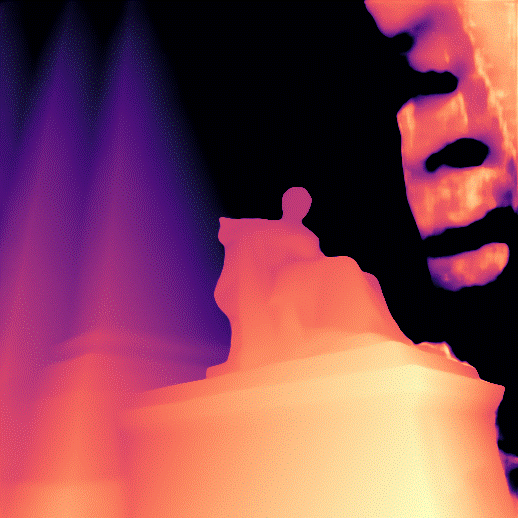} & \includegraphics[width=0.166\textwidth]{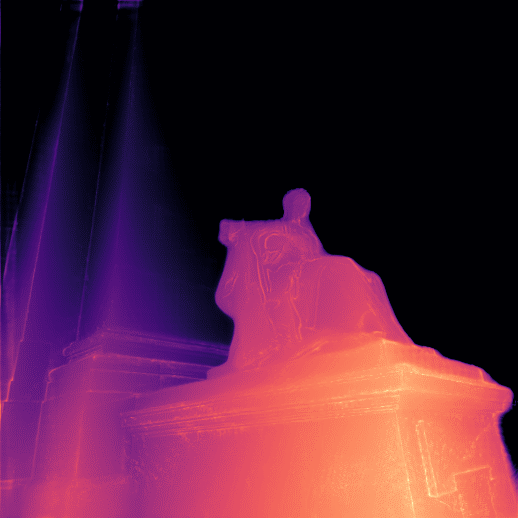} & \includegraphics[width=0.166\textwidth]{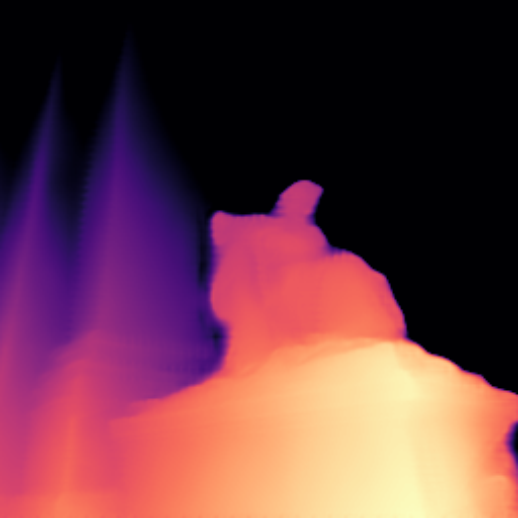} & \includegraphics[width=0.166\textwidth]{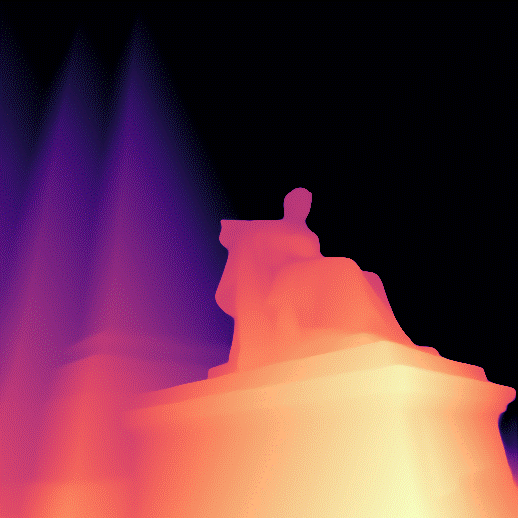}\tabularnewline
\includegraphics[width=0.166\textwidth]{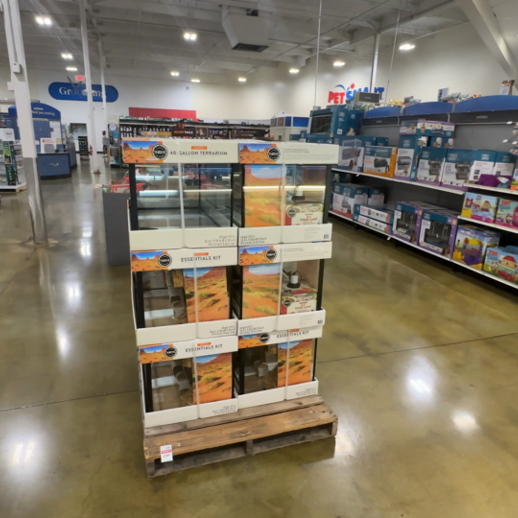} & & \includegraphics[width=0.166\textwidth]{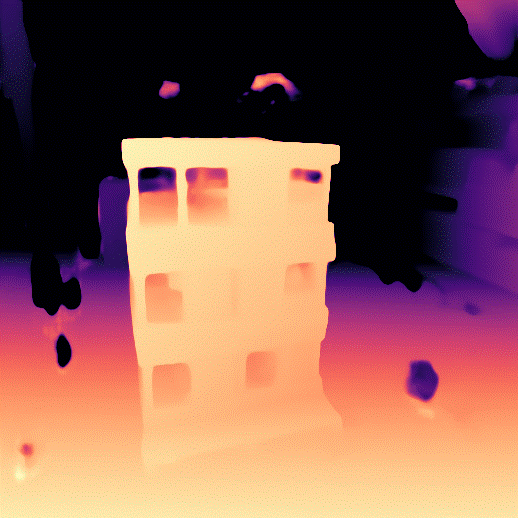} & \includegraphics[width=0.166\textwidth]{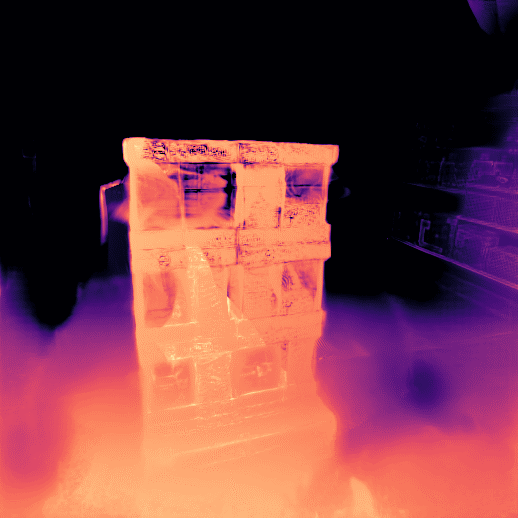} & \includegraphics[width=0.166\textwidth]{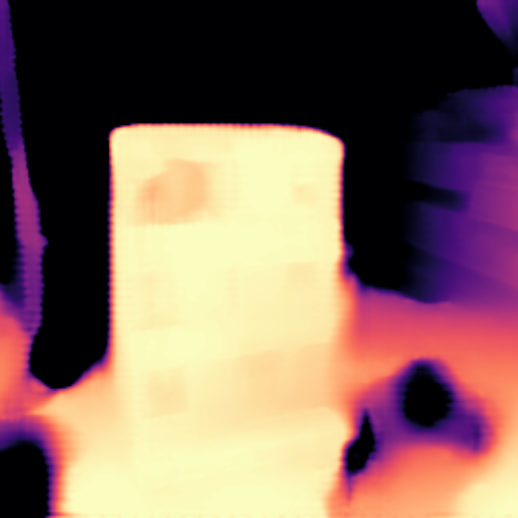} & \includegraphics[width=0.166\textwidth]{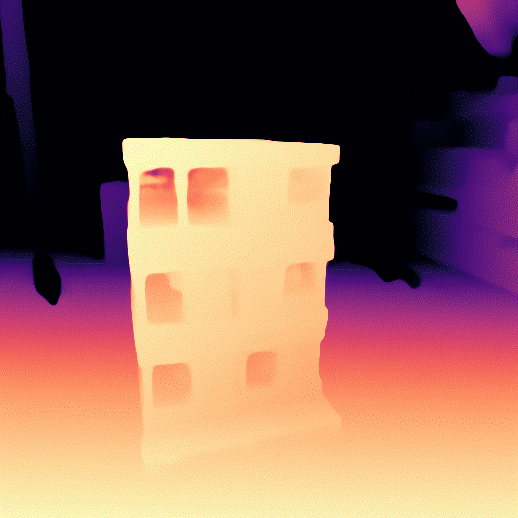}\tabularnewline
\includegraphics[width=0.166\textwidth]{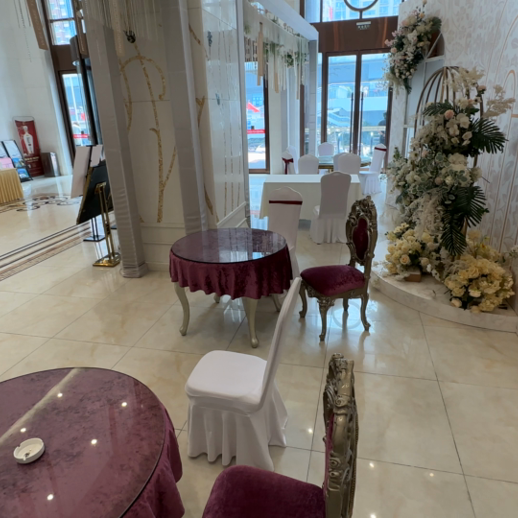} & & \includegraphics[width=0.166\textwidth]{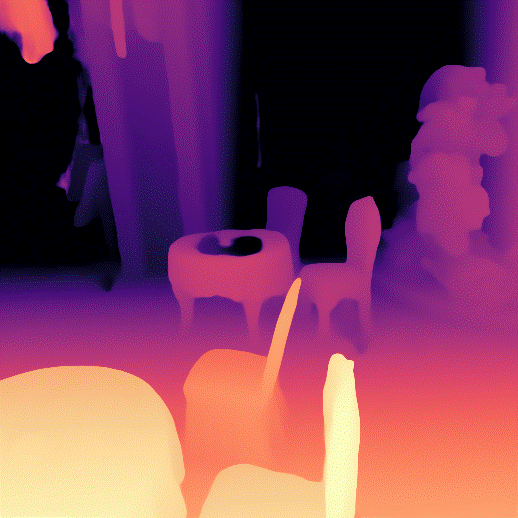} & \includegraphics[width=0.166\textwidth]{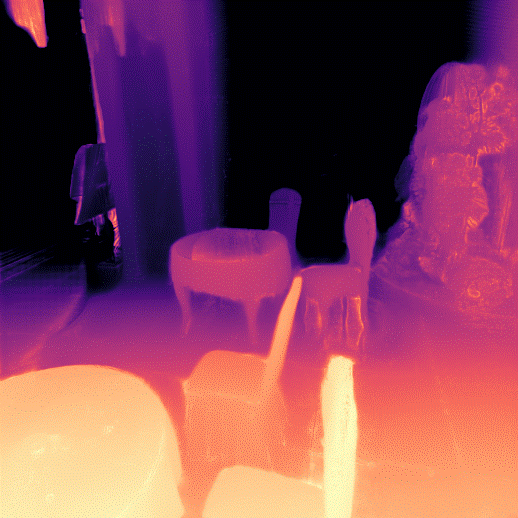} & \includegraphics[width=0.166\textwidth]{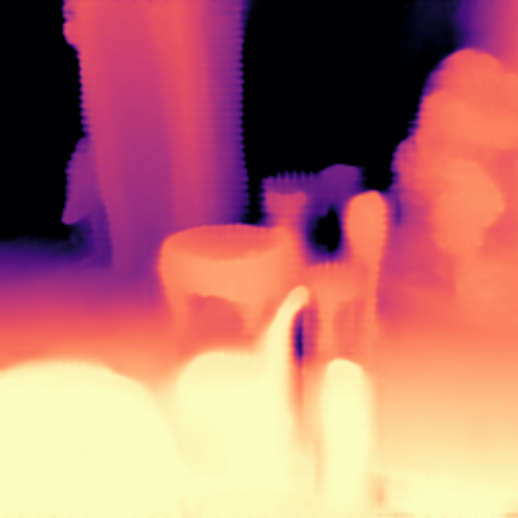} & \includegraphics[width=0.166\textwidth]{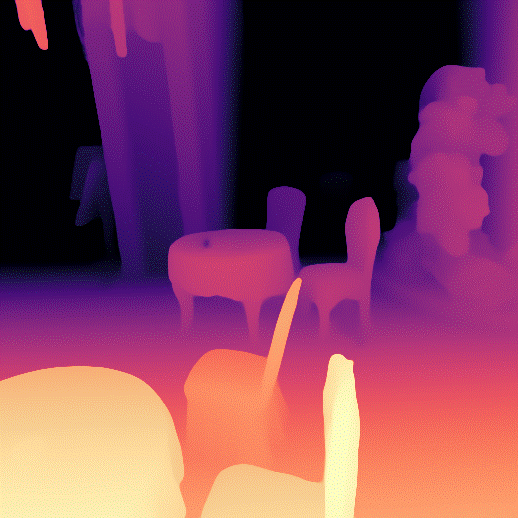}\tabularnewline
\includegraphics[width=0.166\textwidth]{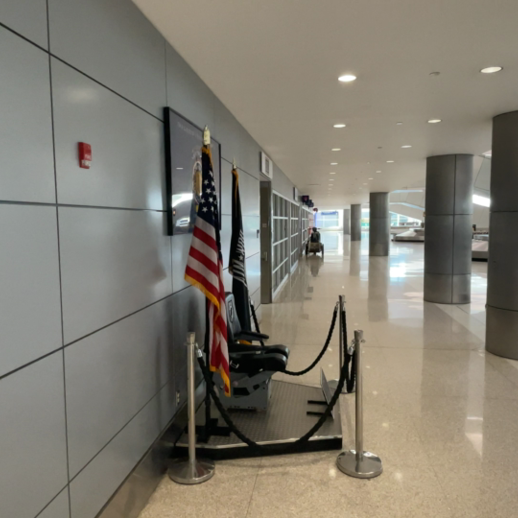} &  & \includegraphics[width=0.166\textwidth]{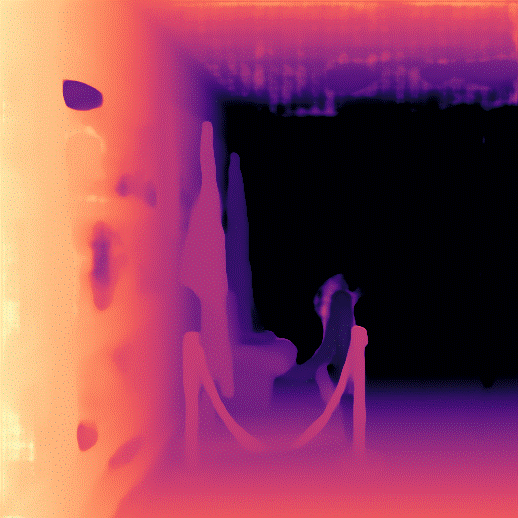} & \includegraphics[width=0.166\textwidth]{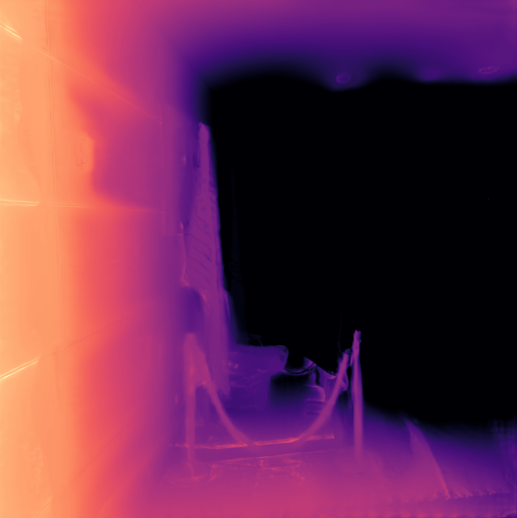} & \includegraphics[width=0.166\textwidth]{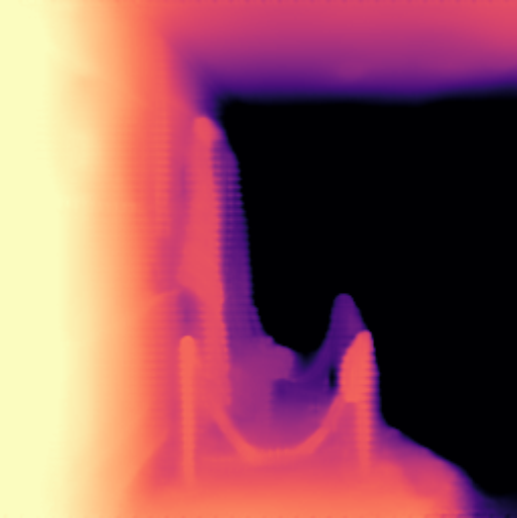} & \includegraphics[width=0.166\textwidth]{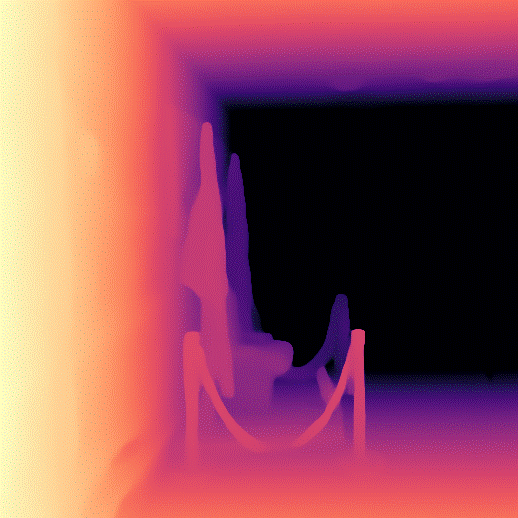}\tabularnewline
\end{tabular}
\par\end{centering}
\caption{Qualitative results of our method compared with several SoTA on depth estimation. }
\label{fig:depth_quali}
\end{figure*}

We present a qualitative evaluation of depth, shown in Fig.~\ref{fig:depth_quali}. The first two rows show examples from Charge, whereas next rows are from DL3DV where ground truth depth is not available. 

First, on Charge, we observe accurate and highly similar depth maps between pre-trained VGGT and our fine-tuned version on this synthetic dataset. The main difference is observed on background areas (see second row, where our method is better aligned with GT for foreground areas but background is predicted too close, degrading metrics). On DL3DV, we observe more insights and failure cases from the pre-trained VGGT model. First, this model is quite sensitive to specularities and create holes on flat but highly reflective surfaces (see rows 3, 5 and 6). Then, during the supervised training of this model, sky pixels were masked out, resulting in close depth estimation for these pixels (row 4), which is not compatible with our rendering task. We also sometimes observe inaccuracy on some flat surfaces (e.g. the ceiling in row 7) without clear reasons. All these failures create geometrically inaccurate models with floating artefacts when we start to train our method. We observe that self-supervised fine-tuning through rendering is able to address theses issues and obtain more accurate depth maps without holes, for both our method and AnySplat~\cite{jiang2025anysplat}, to a lesser degree. We observe that AnySplat depth maps are less accurate than ours, one recurrent artefact being edges appearing where depth is continuous (see row 4). DepthSplat~\cite{xu2024depthsplat} also claims to learn a depth estimation module from rendering but its accuracy is not comparable with VGGT-based models.

\end{document}